\documentclass[review]{elsarticle}

\usepackage{amsmath,amssymb,amsfonts}
\usepackage{lineno,hyperref}
\usepackage{graphicx}
\usepackage{textcomp}
\usepackage{xcolor}
\usepackage{algorithm}
\usepackage{algpseudocode}
\usepackage{svg}
\usepackage{caption}
\usepackage{subcaption}
\usepackage{filecontents}
\usepackage{comment}
\usepackage{booktabs}
\usepackage{colortbl}
\usepackage{graphicx}
\usepackage{supertabular}
\usepackage{tabularx}
\usepackage{array}
\usepackage{multirow}
\usepackage{longtable}
\usepackage{ltablex}
\usepackage{fixme}
\usepackage{lscape}
\usepackage{afterpage}
\usepackage{supertabular}
\usepackage{filecontents}
\usepackage{mathtools}

\DeclarePairedDelimiter\floor{\lfloor}{\rfloor}
\fxsetup{status=draft, theme=color}

\usepackage[a4paper,
            bindingoffset=0.2in,
            left=1in,
            right=1in,
            top=1in,
            bottom=1in,
            footskip=.25in]{geometry}

\journal{Elsavier Journal of \LaTeX\ Templates}

\begin{document}

\begin{frontmatter}


\title{A Three-Stage Algorithm for the Closest String Problem
on Artificial and Real Gene Sequences}

\author[uiasbs]{Alireza Abdi\corref{mycorrespondingauthor}}
\cortext[mycorrespondingauthor]{Corresponding author}
\ead{alirezaabdi@iasbs.ac.ir}

\author[ubanjaluka]{Marko Djukanovic}
\ead{marko.djukanovic@pmf.unibl.org}

\author[uiasbs]{Hesam Tahmasebi Boldaji}
\ead{hesam.tahmasebi@iasbs.ac.ir}

\author[uiasbs]{Hadis Salehi}
\ead{hadith.salehi@iasbs.ac.ir}

\author[ubelgrade]{Aleksandar Kartelj}
\ead{aleksandar.kartelj@matf.bg.ac.rs}

\address[uiasbs]{Department of Computer Science and Information Technology,\\ Institute for Advanced Studies in Basic Sciences (IASBS), Zanjan, Iran}
\address[ubanjaluka]{Mladena Stojanovića 2, 78000, Banja Luka, Republic of Srpska, Bosnia and Herzegovina}
\address[ubelgrade]{Department for Computer Science, Faculty of Mathematics, University of Belgrade, Belgrade, Serbia}

\begin{abstract}
The Closest String Problem is an NP-hard problem that aims to find a string that has the minimum distance from all sequences that belong to the given set of strings. Its applications can be found in coding theory, computational biology, and designing degenerated primers, among others. There are efficient exact algorithms that have reached high-quality solutions for binary sequences. However, there is still room for improvement concerning the quality of solutions over DNA and protein sequences. In this paper, we introduce a three-stage algorithm that comprises the following process: first, we apply a novel alphabet pruning method to reduce the search space for effectively finding promising search regions. Second, a variant of beam search to find a heuristic solution is employed. This method utilizes a newly developed guiding function based on an expected distance heuristic score of partial solutions. Last, we introduce a local search to improve the quality of the solution obtained from the beam search. Furthermore, due to the lack of real-world benchmarks, two real-world datasets are introduced to verify the robustness of the method. The extensive experimental results show that the proposed method outperforms the previous approaches from the literature.
\end{abstract}


\end{frontmatter}

\section{Introduction}\label{intro}
The Closest String (CS) problem involves finding a string that is closest to a given set of strings; the term ``closest'' usually refers to the minimum Hamming distance but may refer to the other sequential measures. CP problem has many applications in coding theory~\cite{li2002}, and computational biology~\cite{lanctot2003}. More specifically, this problem is used in designing drugs, degenerated primers, diagnostic probes, and finding gene clusters~\cite{lanctot2003, macario2012, li1999, ma2008, gramm2002, hufsky2011}.
CS problem is also known as the Hamming Center problem and comes with various versions of different computational complexities. It is known that the CS problem is NP-hard in general~\cite{lanctot2003}, while its decision problem variant is NP-complete~\cite{dor1997}. In the decision version of the CS problem, the aim is to answer a binary question ``Does the input set of string have a closest string with maximum distance of $d$?''. The general version of the CS problem seeks a string that is as close as possible to all given input strings. This paper tackles the general form of the CS problem with an arbitrary set of strings. Let $S = \{s_1, s_2, \cdots, s_n \}$ be the set of strings over a finite alphabet $\Sigma = \{\sigma_1, \sigma_2, \cdots, \sigma_m\}$, where $n$ and $m$ refer to the number of strings and alphabet set cardinality,  respectively. Without loss of generality, we assume that all strings have the same length ($L$). The closest string is a string with length $L$ that has the minimum Hamming distance from strings in $S$.

CS problem has attracted many researchers to solve it exactly as well as heuristically. Gramm et al.~\cite{gramm2003} proposed an exact fixed-parameter algorithm that runs in linear time assuming the Hamming distance is specified and fixed. Meneses et al.~\cite{meneses2004} developed an integer linear programming to solve the CS in general case. This algorithm has been efficient in finding an exact solution in a short time for small and near-moderate instances. However, a huge amount of time is requested for solving larger instances, specifically for random strings with $n > 30$ and $L \geq 800$. In 2011, Liu et al.~\cite{liu2011} introduced an exact algorithm for solving the CS problem for the particular case when $n = 3$ and $|\Sigma| = 2$. Among the exact methods, the ILP model proposed by Meneses et al.~\cite{meneses2004} is the most effective one and extremely efficient for binary strings. 

When it comes to DNA and protein sequences, that are much larger and more complicated to solve, exact methods perform poorly. Therefore, approximation algorithms are widely applied. For example, approximation algorithms with approximation guarantees are proposed~\cite{li2002, lanctot2003, gkasieniec1999} and despite their speed, they showed poor performance in finding near-optimal solutions. (Meta)heuristic methods instead, could find near-optimal solutions in a reasonable time, but they do not guarantee optimality in general. Nevertheless, the heuristic methods outperform the previously mentioned methods when applied to large DNA and protein sequences. 
In 2005, Liu et al.~\cite{liu2005} suggested using genetic and simulated annealing algorithms to solve the CS problem. Moreover, they implemented the parallel version of the algorithm and showed that the parallel genetic algorithm is superior when compared to the others. Later, Gomes et al.~\cite{gomes2005, gomes2008} proposed a simple heuristic along with its parallel version.
In 2008, Liu et al.~\cite{liu2008} proposed a hybridization of genetic and simulated annealing algorithms and combined their metrics to solve the CS problem. However, their algorithm is only tested for binary strings.
Later in 2010, Faro and Pappalardo~\cite{faro2010} introduced an algorithm based on the Ant Colony Optimization method which excelled over all former approaches. 
Mousavi and Esfahani~\cite{mousavi2012} suggested using the greedy randomized adaptive search procedure (GRASP) algorithm followed by a new heuristic function inspired by probability theory to solve the general case of the CS problem.
In 2011, Mousavi~\cite{mousavi2010} utilized a beam search as a part of a hybrid meta-heuristic and also, while modifying the heuristic function from~\cite{mousavi2012} to get a more enhanced one.
Liu et al.~\cite{liu2011} in their other work from 2011, built on their heuristic function from~\cite{liu2004} to design an effective local search.
In 2014, Pappalardo et al.~\cite{pappalardo2014} proposed a hybrid of a greedy algorithm and simulated annealing (GWSA) to solve the CS problem. They first generated an initial solution by a greedy walk algorithm, next, they fed the output of the greedy walk algorithm to a simulated annealing algorithm. GWSA excelled over all former approaches but delivered high running time.  
Recently, Xu and Perkins~\cite{xu2022} proposed using Wave Function Collapse (WFC) to solve the CS problem, claiming it outperforms all other approaches. However, they did not compare their method with the GWSA. The WFC is a lightweight method with very low running time. First, they constructed a frequency table based on the repetition of characters in strings. Second, they identified the string with the maximum Hamming distance. Finally, according to the frequency table, they assigned a character to the solution. They repeated the second and the last steps until a solution of complete length was obtained. This method is particularly suitable for solving the CS problem when dealing with protein sequences. In summary, the WFC is considered the state-of-the-art heuristic method for solving the general form of the closest string problem.

Contributions of this paper are as follows.
\begin{itemize}
    \item A \emph{Three-Stage Algorithm} (TSA) to address the CS problem is designed. In the first stage of the algorithm, a novel pruning method that enhances both the speed and quality of the algorithm is performed. The next phase employs a time-restricted variant of beam search on the restricted search space. A new heuristic function based on expected distance score to guide the beam search is designed. In the last stage, an efficient local search is executed to refine the solution obtained by the beam search. 
  \item We have created two real-world datasets and discussed the robustness of the approaches. The first one consists of the well-known TP53 gene nucleotide, and the second dataset comprises the nucleotides of two segments of a common flu variant across different hosts to check their conservation levels.
  \item Extensive experimental evaluation shows that our proposed algorithm (TSA) outperforms the state-of-the-art approaches over all five benchmark sets. 

\end{itemize}

The rest of this paper is structured as follows. In Section~\ref{basic}, we provide several definitions and formulations required for the subsequent sections. Following that, we introduce details of our algorithm in Section~\ref{pro}. Section~\ref{exp} discusses the experimental results along with statistical reports. Finally, Section~\ref{conc} concludes the paper by additionally providing some outlines for future work.

\section{Basic Definitions and Preliminaries}\label{basic}
This section provides the fundamentals and the definitions essential for understanding the major core of the paper. 

Given is a set $S = \{s_1, s_2, \ldots, s_n \}$ of input strings over a finite alphabet \sloppy $\Sigma = \{\sigma_1, \sigma_2, \ldots, \sigma_m \}$, where $n$ and $m$ refer to the numbers of input strings and alphabet cardinality, respectively. Assume that all strings in $S$ have the same length $L$. Let $s_i^j$ denotes the $j^{th}$ character of $i^{th}$ string in $S$, while $s_i^1$ denotes the leading character of string $s_i$.  The notion of level $l$ ($1 \leq l \leq L$) takes all  $l^{th}$ characters over input strings in $S$. For two integer values $1 \leq i \leq j \leq |s|$, by $s[i, j]$ we denote the continuous part of string $s$ (sub-string) that starts with the character at position $i$ and ends with the character at position $j$. 
The frequency of a character ($\sigma$) at level $l$ is denoted by $f_l(\sigma)$, i.e. the number of occurrences of $\sigma$ at level $l$.
Let $S[l]=\{ s_i[1, l] \mid i=1, \ldots, n\}$ be the set of prefix strings of each input string w.r.t. position $l \in \{1, \ldots, L \}$.

In the closest string problem, the Hamming distance ($hd$) represents the distance between strings and the proposed solution. It is computed by counting the differing positions between each string ($s_i$) and the solution, with the largest of these distances representing the solution’s overall Hamming distance. Symbolically, for any two strings ($s_1$) and ($s_2$) of equal length ($L$), the Hamming distance is defined as in Eq.\ref{eq:hamming distance}.
\begin{equation}
    \label{eq:hamming distance}
    \begin{split}
        hd(s_1, s_2) = \sum_{i = 1}^{l} \delta(s_1^i, s_2^i) \\
        \delta(ch_1, ch_2) \leftarrow
        \begin{cases}
            1 & ch_1 \neq ch_2 \\
            0 & ch_1 = ch_2
        \end{cases}
    \end{split}
\end{equation}

To facilitate the explanation of our expected distance heuristic in Section~\ref{ex heuristic}, let $hd^c(s_1,s_2):=\sum_{i=1}^l (1-\delta(s_1^i, s_2^i)$, denote the complementary score of the Hamming distance by $hd(s_1, s_2)$, which refers to the number of equal characters between the two strings at the same positions. Note that $hd(s_1, s_2)=L-hd^c(s_1, s_2).$ 
In essence, the hamming distance between string $s$ and a set of input string $S$ is calculated by
\begin{equation}
    \label{eq:hamming_string_set}
    hd(s, S) = \max \{ hd(s, s_i) \mid i=1, \ldots, n \}. 
\end{equation}

Note that since we solve the CS problem in a constructive manner where our solutions are partial, we explain the hamming distance calculation for a partial solution $s^p$ with length of $|s^p|=l_p$. That is, the hamming distances is calculated applying Eq.~(\ref{eq:hamming_string_set}) as the score $hd(s^p, S[l_p])$. 
One can easily see  that $hd^c(s^p)$ is equal to $l_p - hd(s^p, S[l_p])$.

Beam Search (BS) is a heuristic tree-search algorithm that works in a limited breadth-first manner. It was proposed by Hayes et al.~\cite{hayes1976} and applied in the context of the speech understanding problem.  The BS and its variants are widely applied to solve various optimization problems, e.g., scheduling and machine translation~\cite{ow1988, freitag2017}.   In the context of string problems from bioinformatics, it has been proven as one of the most efficient techniques; for example, it is successfully employed to solve the prominent Longest Common Subsequence problem~\cite{blum2009}. A specific number ($\beta>0$) of nodes based on a heuristic evaluation is selected at each level of the search to be further processed.  BS uses parameter $\beta$ to provide the trade-off between completeness and greediness. As the value of $\beta$ increases, the runtime of the algorithm and the chance of finding the optimal answer gets higher and vice versa. For the CS problem, BS starts with an empty solution and constructs solutions by adding letters from $\Sigma$ to the most promising $\beta$ nodes, which correspond to (actions of) appending these letters to the end of respective partial solutions,  until a complete node is reached. Complete nodes are those whose corresponding partial solutions cannot be further expanded. In the context of CS problems, complete nodes are solutions of length $L$.  

For the sake of clarity, consider the \sloppy $S = \{ s_1 = \texttt{abaaabbaba}, s_2 = \texttt{abababaabb} \}$ and $\Sigma = \{\texttt{a}, \texttt{b}\}$. Here, $n$, $L$, and $m$ are 2, 10, and 2, respectively. For a partial solution $x = \texttt{ababa}$ with length $l_p = 5$, the hamming distance is calculated based on the first 5 characters of all strings i.e., $s_1 = \texttt{abaaa}$, and $s_2 = \texttt{ababa}$, which $hd(x, s_1) = 1$, and $hd(x, s_2) = 0$. Also, $hd^c(x, s_1) = 4$, and $hd^c(x, s_2) = 5$. In addition, $S^4_2$ denotes the fourth character of the second string which is $\texttt{b}$. Furthermore, the frequency of character \texttt{a} and \texttt{b} at level 1 ($f_1(\texttt{a})$ and $f_1(\texttt{b})$) is 2 and 0, respectively.

\section{The Proposed Method}\label{pro}

In this section, the \emph{Three-Stage Algorithm} (TSA) is designed to address the CS problem, pseudocoded by Algorithm~\ref{alg:csimp}. Initially, the algorithm pre-processes the set of effective actions at each level of BS employing the pruning mechanism, which serves to diminish the search space. Subsequently, it employs the expected distance heuristic function to guide the Time-Restricted Beam Search (TRBS)~\cite{nikolic2021} towards a solution of reasonable quality. Lastly, the obtained solution is passed to the new local search algorithm to further refine its quality. \\ To summarize, the methodology applies the three core stages: ($i$)  the novel pruning method described in Section~\ref{pruning section} to support effective decision-making; ($ii$) the TRBS algorithm specifically tailored for the CS problem provided in Section~\ref{TRBS} whose the search is guided by the new expected distance heuristic, supported by the associated tie-breaking strategy; see Section~\ref{ex heuristic}; ($iii$) a new local search algorithm detailed in Section~\ref{local search}.

\begin{algorithm}
\caption{TSA for the CS problem}\label{alg:csimp}
\begin{algorithmic}[1]
\State $\textbf{Input: } S, \beta_t, \beta, t_{max}$
\State \textbf{Initialization}: $l = 0, B = \{$`'$\}, C = \{\}$
\State $\Sigma_P =$ rankIdentify$(S, \Sigma, \beta_t)$
\While{$l < L$} 
    \For{each $v$ in $B$}
        \State $C \leftarrow C \cup expandNodes(v, \Sigma_P)$
    \EndFor
    \For{each $v$ in $C$}
        \State $v.h \leftarrow$ exHeuristic($v$)
    \EndFor
    \If{tie occurred} 
        \For{each $v$ in $C$}
            \State $v.h \leftarrow$ varianceHeuristic($v$)
        \EndFor
    \EndIf
    \State $B \leftarrow$ keepBestNodes$(C, \beta)$
    \State $\beta \leftarrow$ adjustBeta$(\beta, l, t_{max})$
    \State $l \leftarrow l + 1$
\EndWhile
\State $cur\_best\_sol \leftarrow$ bestSolution$(B)$
\State $imp\_best\_sol \leftarrow$ localSearch$(cur\_best\_sol, S)$
\State \Return $imp\_best\_sol$
\end{algorithmic}
\end{algorithm}

\subsection{A Novel Pruning Method for the Closest String Problem}\label{pruning section}

Due to the NP-hardness of the CS problem, the state space grows exponentially with the instance size.
The following question arises: Should we regard all members of $\Sigma$ as candidates at level $l \in \{1, \ldots, L\} $ for a solution? To answer this question, we have analyzed the exact solution for 50 instances with a length of 100 and it turned out that in most cases, the selected character was from a specific subset of characters. Hence, we decided to design a pruning method to determine those specific subsets of alphabets and consequently, improve the quality of solutions and running time of the BS approach. To describe the pruning method, let us first define the notion of `rank 1'  (R1) and `rank 2' (R2). For the following list of tuples $[(A, 10), (B, 12), (C, 10), (D, 7), (E, 1), (F, 6), (G, 9)]$ where the first coordinate associates a character and the second coordinate refers to the number of occurrences of that letter, the R1 assigns the set of characters with the highest second coordinate, which here would be $B$. The R2 contains the characters with the first and second highest assigned values; in this example, the first highest value is given to $B$, and the second highest value (10) is assigned to $A$ and $C$. As the values for both of them are the same, both are assigned to R2. Thus, the R2 contains the three characters, $B$, $A$, and $C$. \\

To clarify on a concrete instance, consider Table~\ref{tab:pruning}, $S = \{s_1, s_2, \ldots, s_8\}$ whose the length of input strings are \textit{10}. Now, we should determine the R1 and R2 for each $l \in  \{1, \ldots, L \}$. To do this, we use the frequency of characters at each level $l$. For example at the fixed level $l = 6$, the frequency of characters is as follows, $f_6(X) = 2, f_6(K) = 3, f_6(E) = 2, f_6(C) = 1$, and the frequency of the rest of $\Sigma$ members are $0$. According to the introduced notation, the most frequent characters are assigned R1, which is $K$. For identifying the R2 set, we add the first and second most frequent characters here; thus, $K$ is the first, and since the $X$ and $E$'s frequencies match, both of them are considered as the second most frequent characters. Hence, the R1 set for ($l = 6$) is $\{K\}$, and the R2 set for ($l = 6$) is $\{K, X, E\}$. It is notable that after obtaining ranks, these pairs (R1, R2) are stored in a list of alphabets for each $l$, called the ranked alphabet ($\Sigma_P$) which are subsequently passed to the BS procedure. 

For example, in Table~\ref{tab:pruning}, consider we want to use R2, so, our solution's first character could only be `M' or `C'. During the preprocessing phase and before obtaining the CS for a set of input strings, it is essential to identify the proper rank that will be leveraged. The choice between R1 and R2 should be guided by the inherent characteristics of the input strings in question, as this decision is dependent on specific conditions. 
A simple way~\cite{tabataba2012} to determine the most appropriate rank for a given set of strings is to initially execute the BS with a small (trial) $\beta_t$ value using both ranks. 
Subsequently, the algorithm runs with a larger (actual) $\beta$ value using the rank that yielded a smaller hamming distance. When the hamming distances are equal for both ranks, the option is selected randomly. This pruning method could be used in all methods that tried to solve the CS problem and it is not limited to our approach.

\begin{table}[]
\caption{This table shows the R1 and R2 sets for the instance consisting of 8 input strings, each with length 10.}
\label{tab:pruning}
\resizebox{\columnwidth}{!}{
\begin{tabular}{|c|c|c|c|c|c|c|c|c|c|c|}
\hline
\multicolumn{1}{|c|}{} & \multicolumn{1}{c|}{$l = 1$} & \multicolumn{1}{c|}{$l = 2$} & \multicolumn{1}{c|}{$l = 3$} & \multicolumn{1}{c|}{$l = 4$} & \multicolumn{1}{c|}{$l = 5$} & \multicolumn{1}{c|}{$l = 6$} & \multicolumn{1}{c|}{$l = 7$} & \multicolumn{1}{c|}{$l = 8$} & \multicolumn{1}{c|}{$l = 9$} & \multicolumn{1}{c|}{$l = 10$} \\ \hline
$s_1 $                  & M                          & K                          & D                          & L                          & E                          & X                          & H                          & X                          & A                          & L                           \\ \hline
$s_2$                   & X                          & X                          & T                          & D                          & Y                          & K                          & N                          & S                          & M                          & I                           \\ \hline
$s_3 $                  & M                          & F                          & W                          & H                          & T                          & E                          & H                          & Y                          & H                          & I                           \\ \hline
$s_4$                   & D                          & H                          & G                          & C                          & P                          & C                          & V                          & G                          & H                          & W                           \\ \hline
$s_5$                   & C                          & Y                          & L                          & A                          & T                          & K                          & Q                          & I                          & I                          & X                           \\ \hline
$s_6$                   & M                          & A                          & M                          & S                          & S                          & X                          & N                          & G                          & H                          & I                           \\ \hline
$s_7$                   & Q                          & K                          & S                          & C                          & Y                          & K                          & L                          & S                          & V                          & Q                           \\ \hline
$s_8$                   & C                          & H                          & W                          & D                          & T                          & E                          & H                          & S                          & H                          & W                           \\ \hline \hline
\multicolumn{1}{|c|}{R1} & \multicolumn{1}{c|}{M}     & \multicolumn{1}{c|}{H}     & \multicolumn{1}{c|}{W}                   & \multicolumn{1}{c|}{D, C}             & \multicolumn{1}{c|}{T}     & \multicolumn{1}{c|}{K}       & \multicolumn{1}{c|}{H}     & \multicolumn{1}{c|}{S}     & \multicolumn{1}{c|}{H}             & I                           \\ \hline
\multicolumn{1}{|c|}{R2} & \multicolumn{1}{c|}{M, C}  & \multicolumn{1}{c|}{H, K}  & \multicolumn{1}{c|}{W, D, T, G, L, M, S} & \multicolumn{1}{c|}{D, C, L, H, C, A} & \multicolumn{1}{c|}{T, Y}  & \multicolumn{1}{c|}{K, X, E} & \multicolumn{1}{c|}{H, N}  & \multicolumn{1}{c|}{S, G}  & \multicolumn{1}{c|}{H, A, M, I, V} & I, W  \\ \hline                
\end{tabular}}
\end{table}

\subsection{Time Restricted Beam Search for the  CSP}\label{TRBS}
To explore the search space of the CS problem effectively, a time-restricted version of the BS approach is employed. This version, therefore called the \emph{Time-Restricted Beam Search} (TRBS)~\cite{nikolic2021},  has proven to be more robust than the basic BS variant with a provided good trade-off between greediness and completeness~\cite{nikolic2021}. Moreover, it addresses the problem of adjusting the value of $\beta$ parameter dynamically within the allowed execution time ($t_{max}$). Note that this is not the characteristic of the basic beam search. At each major iteration of the algorithm, TRBS identifies $\beta$ value depending on the remaining runtime and the value of $\beta$ used in the previous iteration. 
The details of computing $\beta$ at iteration  $l$ are given by Eq.~\ref{eq:TRBS}.

\begin{equation}
    \label{eq:TRBS}
    \beta \leftarrow
    \begin{cases}
             \floor*{\beta\cdot1.1} & \text{if $t_{rem} / \overline{t_{rem}}  \geq 1.1$;}\\
             \min(\floor*{\beta / 1.1}, 150) & \text{if $t_{rem} / \overline{t_{rem}}  \leq 0.9$;} \\
             \beta & \text{otherwise.}
    \end{cases}
\end{equation}

Here, $t_{rem}$ denotes the remaining time to reach the $t_{max}$ and $\overline{t_{rem}}$ is the expected remaining time which for CS problem is equal to $t_{iter} \cdot (L - l)$. In other words, $\overline{t_{rem}}$ estimates the remaining time by multiplying the last iteration time ($t_{iter}$) with the remaining number of steps to achieve a leaf node. In the context of the CS problem, since the length of each leaf node (a solution) is equal to $L$ supposing the BS performs at the current level $l$, the remaining number of BS iterations would be ($L - l$).

\subsection{An Expected Distance Heuristic for CSP}\label{ex heuristic}

This section is devoted to a constriction of the expected distance heuristic (EX) to evaluate nodes of the TRBS. Additionally, an effective way of breaking ties between nodes with the same expected distance score values is proposed.

First, we explain the term of an expected solution to the CS problem. For each level $l \in \{1, \ldots, L\}$, we determine the most frequent letter across strings from $S$. Assume that the information is kept in the (sequential) structure \texttt{expected\_sol}, where $\texttt{expected\_sol}[l]$ stores the information about the most frequent letter at level $l$ across all input strings in $S$.

Suppose that $s$ is a partial solution of a fixed length $l<L$ which has to be evaluated.  Based on the expected solution string 
\texttt{expected\_sol}, let us approximate the expected distance from this partial solution to a complete solution. To do so, for each string $s_j$ and the expected string \texttt{expected\_sol}, the number of positional matchings between each suffix of \texttt{expected\_sol} and $s_j$ of the same size in both strings are calculated. For example, if $\texttt{expected\_sol}=\texttt{abbc}$ and $s_j=\texttt{acbe}$, the scoring vector is obtained $[0, 1, 1, 2]$ by first comparing characters at position 4 (does not match), then at position 3 (which matches, yielding a score of 1), etc.  
Reversely,  the vector $score_j=[2, 1, 1, 0]$ is constructed for each $j=1,\ldots, n$. This means, that $score_j[r]$ immediately restores the number of characters between the $r$-length suffix of $\texttt{expected\_sol}$ and the $r$-length suffix of $s_j$ that match at the same position (index) between these two. These structures are therefore preprocessed before entering the main loop of the BS. The score of solution $s$ is based on the expected number of matched letters obtained by taking into account two scores for each input string $s_i$: ($i$) the number of positions between the prefix of $s_i$ of length $l$ and solution $s$ for which respective characters match each other, and ($ii$) estimated number of positions between the suffixes of \texttt{expected\_sol} and $s_i$ with the starting position $l+1$ whose characters match each other. In essence, the final score is based on the most probable suffix solution  \texttt{expected\_sol} for the remaining parts (suffixes) of input strings relevant for extending partial solution $s$. Symbolically, for a solution $x$, $l=|x|$, the following vector of dimension $n$ is calculated:
\begin{align}\label{eq:ex-long}
    fitness\_vector(x)&=hd^c(x, S[\textbf{l}]=\{s_1[1, l], \ldots, s_n[1, l]\})\\ &+(score[l+1, L]), \nonumber
\end{align}
where $hd^c(x, S)=( hd^c(x,s_1), \ldots, hd^c(x, s_n))$, and the plus operator on vectors is applied coordinate-by-coordinate. Now, an estimated expected distance heuristic score for solution $s$ and the set of input strings $S$ is given by
\begin{equation}\label{eq:ex-csp}
    EX(x)=\min_{i=1, \ldots, n} fitness\_vector(x).
\end{equation}

Note that incorporating the minimum in Eq.~(\ref{eq:ex-csp}) has been found the most suitable as utilizing the maximum or the mean value has been shown as too optimistic estimators, yielding weaker performing search guidance. As an important detail, the first term in Eq.~(\ref{eq:ex-long}) is calculated incrementally, updating the score from the parent node of the node associated with partial solution $x$. This operation is executed in $O(n)$ time. 
The second term, using the preprocessed data structures \textit{score}, requires $O(n)$ time. Therefore, one can calculate  Eq.~(\ref{eq:ex-csp}) efficiently, in a linear $O(n)$ time.  Note that nodes with larger \emph{EX} values are preferred.  

In the case of ties among nodes at the same level (i.e., having the same EX($\cdot$) score for each node), we break them by utilizing another heuristic rule based on the variance of hamming distances between solution $x$ and corresponding prefixes of input strings; a smaller variance is preferred. For a partial solution $x$, the variance is calculated based on Eq.~\ref{eq:var}.

\begin{equation}  
    \label{eq:var}
    Var(x) = \frac{\sum_{i = 1}^{n} (hd(x, s_i) - \overline{hd} )^2 }{n - 1},
\end{equation}
Where $\overline{hd}$ is the mean of hamming distances between $x$ and all strings, i.e. $\overline{hd}=\frac{\sum_{i=1}^{n}hd(x, s_i)}{n}$.

\subsection{A New Local Search for CS problem}\label{local search}

In this section, we present a new local search algorithm that starts with the initial solution obtained by executing the BS algorithm and intends to improve its quality by performing certain character modifications. The main idea of this procedure is to flatten peak strings with the largest Hamming distances in a way that minimizes the impact on other strings. In more detail, we refer to Algorithm~\ref{alg:ls}: the preliminary step involves identifying the critical strings that possess the maximal distances between the incumbent solution  $best\_sol$ and all input strings from $S$. In the case where multiple critical strings exist, all of them are taken into account. Subsequently, the character that appears most frequently (the character with the highest $f_l(\cdot)$, $1 \leq l \leq L$) within these critical strings is recognized along with its position of appearance, denoted by \emph{index}. In case of multiple characters with the same frequency, all of them are considered. We store all these pairs of indices and characters in a list \emph{All\_Pairs\_Indices\_Chars} which is then shuffled to prevent the search from any bias. We iterate through the list by considering each pair $(index, char)$ to replace \emph{char} into a temporary solution $temp_{sol}$ at the position \emph{index}. If this modification leads to a hamming distance that is either reduced or equivalent to that of the prior optimal solution, it is declared for the new best solution, i.e. $best_{sol}=temp_{sol}$ and the next iteration of LS has been performed. If none of the pairs $(index, char)$  yield a better or equivalent Hamming distance, the algorithm terminates by returning  $best\_sol$ even before the time limit is reached. Otherwise, the above procedure repeats until a certain time limit is reached. 

Consider the following example of a major iteration of the local search algorithm. Let $S=\{s_1 = \texttt{CAGTG},$ $s_2 = \texttt{CGATA},$ $s_3 = \texttt{GATCA},$ $s_4 = \texttt{CTACG}\}$. Given the initial solution $best\_sol = \texttt{GAACG}$, we compute the hamming distances as follows: $hd(best_{sol}, s_1) = 3$, $hd(best_{sol}, s_2) = 4$, $hd(best_{sol}, s_3) = 2$, $hd(best_{sol}, s_4) = 2$. At first, the critical strings with maximum hamming distances must be identified, which in this case is $s_2$ with a distance of $4$. Now, to reduce the hamming distance while to minimally impact the other strings, we analyze the character frequency in $s_2$, yielding: $f_1(s_2^1: \texttt{C}) = 3$, $f_2(s_2^2: \texttt{G}) = 1$, $f_3(s_2^3: \texttt{A}) = 2$, $f_4(s_2^4: \texttt{T}) = 2$, $f_5(s_2^5: \texttt{A}) = 2$. As it is clear, the most frequent character of $s_2$ is $\texttt{C}$ at level $1$ with a frequency of $3$. By placing the character $\texttt{C}$ at the first position of $temp_{sol}=best_{sol}$, we derive a new (temporary) solution $temp_{sol}= \texttt{CAACG}$ resulting in the updated hamming distances: $hd(temp_{sol}, s_1) = 2$, $hd(temp_{sol}, s_2) = 3$, $hd(temp_{sol}, s_3) = 3$, $hd(temp_{sol}, s_4) = 1$. In this way, the hamming distance of the modified solution is reduced (from $4$) to $3$ and,  therefore, $best_{sol}$ is set to $temp_{sol}$.

\begin{algorithm}
\caption{Local Search}\label{alg:ls}
\begin{algorithmic}[1]
\State $\textbf{Input: } init\_sol, S$
\State $best\_sol = temp\_sol \gets init\_sol$
\State $max\_hm\_dist \gets  hd (best\_sol, S)$

\While{time limit is not reached}
    \State $Terminate \gets True$
    \State $critical\_strings \gets$ \texttt{findCritStrings}$(best\_sol, S, max\_hm\_dist)$
    \State $All\_Pairs\_Indices\_Chars \gets $
    \texttt{mostFreqChar}$(critical\_strings)$ 
    \State  $All\_Pairs\_Indices\_Chars \gets  Shuffle(All\_Pairs\_Indices\_Chars)$
    \For{each $index$, $char$ in $All\_Pairs\_Indices\_Chars$}
    \State $temp\_sol[index] \gets char$
     \State $temp\_hm\_distance \gets hd(temp\_sol)$ \# calculated incrementally
     \If{ $temp\_hm\_distance \leq max\_hm\_dist$}
        \State $best\_sol \gets temp\_sol$
        \State $max\_hm\_dist \gets temp\_hm\_distance$
        \State $Terminate \gets False$
        \State $\textbf{break}$ \#accept equally good or better sol. and move on
     \Else
         \State $temp\_sol \gets best\_sol$ 
     \EndIf
     
     \EndFor
     \If{$Terminate = True$} \#  a local optimum is reached
        \State \textbf{return} $best\_sol$
     \EndIf
\EndWhile
\State \Return $best\_sol$
\end{algorithmic}
\end{algorithm}

\section{Experimental Results and Statistical Analysis}\label{exp}
In this section, we compare our method with state-of-the-art approaches to the CS problem from the literature. To the best of our knowledge, ILP, GWSA, and WFC have reported the best results on the known CS problem instances from the literature. Since the ILP model designed by Meneses et al.~\cite{meneses2004} is extremely strong and more effective than any heuristic approach for binary strings, clearly state-of-the-art method there, we compare our TSA on DNA and protein strings against the ILP, GWSA, and WFC methods. To facilitate this comparison, we first detail the benchmark datasets used for the comparison purposes as given in Section~\ref{datasets}. We then describe the implementation details and parameters' tuning in Section~\ref{implementation}. Numerical comparison of the methods over five benchmark datasets is presented in Section~\ref{comparison}. Section~\ref{time comparison} discusses the run-time analysis of the approaches, and finally, we analyze the methods from a statistical perspective in Section~\ref{statistical}.

\subsection{Benchmark Datasets}\label{datasets}

In the CS problem literature, we found a big confusion with the employed random instances as each work produces its own randomly generated dataset for testing purposes. We emphasize that the datasets from most of the authors of the existing work are requested. However, none of them responded to our queries. Thus, we were forced to produce another uniformly generated dataset that consists of a wide range of strings. Two datasets called \textit{Alpha-4} and \textit{Alpha-20} are generated for comparison purposes which correspond to artificial gene sequences representing DNA and protein sequences. In order to resolve the aforementioned confusion with instances,  our instances can be found at the publicly available GitHub repository at the link \url{https://github.com/Hesam1991/Three-Stage-Algorithm-for-the-Closest-String-Problem}.
Both \textit{Alpha-4} and \textit{Alpha-20} consist of strings with $n \in \{10, 15, 20, 25, 40, 60, 80, 100\}$ and $L \in \{50, 100, 150, 200, 250, 300, 350, 400, 800, 1000, 1500\}$ to test the algorithm in different situations. This gives us $88 \cdot 2=176$ random instances.
Furthermore, the \textit{McClure} is a small real-world protein dataset that is widely used to test the performance of the algorithms. It consists of six instances with $n \in \{6, 10, 12\}$ and $L \in \{98, 100, 141 \}$. We requested the \textit{McClure} dataset from the authors of the original paper, but the instances could not  be provided to us. After an intensive web search, this dataset is found but, unfortunately, one of these instances was corrupted. However, we report the results of the algorithms on the remaining five instances. 

The \textit{McClure} is a small dataset that includes protein sequences and realistic instances are much larger, we introduce a real-world DNA dataset from the TP53 gene which is known as the guardian of the genome. TP53 plays a crucial role in suppressing the tumor cells in living creatures. When the TP53 gene is damaged, it cannot repair itself and consequently, it cannot prevent cancer. The sequence of the TP53 gene differs for distinct living creatures, but its functionality is the same in all of them. Understanding the conservation levels of these genes can help to predict a species’ cancer resistance~\cite{nair2022}. We have collected the TP53 gene of 27 animals including humans, chimpanzees, elephants, etc. from NCBI and randomly put them together as a dataset. This dataset consists of nine instances with $n \in \{5, 10, 27\}$, and $L \in \{500, 1000, 2000, 3000 \}$. 

One application of the CS problem consists in designing degenerated primers for detecting viral diseases. The flu disease type \textit{A} sub-type \textit{H1N1} is one of the common strains affecting both humans and animals. This type of flu mutates frequently and consists of eight segments. In some segments, the mutation rate is high, while in others, it appears to be low. Mainly, two of these segments—4 (HA) and 6 (NA)—are used in designing degenerated primers~\cite{Greninger2010}. Thus, we decided to review these two segments to check the conservation level across different hosts. To do this, we collected the nucleotide sequences of the mentioned flu disease from various hosts and years via NCBI and compiled them into a dataset with ten instances (five for HA and five for NA), each containing 20 sequences ranging from 1000 to 1700 nucleotides. According to the obtained distances in Table~\ref{tab:flu}, the conservation levels in both considered segments are close to each other, but segment 4 (HA) is slightly more conserved (the difference is at about 2\%) than segment 6 (NA).

\subsection{Implementation Details and Parameters Tuning}\label{implementation}

In this section, we describe the way we implemented and set the parameters of the TSA method as well as our three competitors (ILP, GWSA, and WFC). We requested the authors of the mentioned methods for their original implementations. The authors of the WFC algorithm provided us with their Python code; however,  any response has not been received by the authors of the ILP, and GWSA. Therefore, we carefully re-implemented these two approaches. Our re-implementations are freely accessible via previously mentioned GitHub repository. 
All implementations are coded in Python 3.8. To solve the ILP model, CPLEX 12.8 has been executed under default settings. All experiments were executed in single-threaded mode on a machine equipped with an Intel Core i7-4702MQ processor running at 2.2 GHz and using 6 GB of memory.
ILP and WFC have no parameters to be tuned. For the GWSA method, we used the parameters as mentioned in~\cite{pappalardo2014}. For tuning the TSA parameters, we have used grid search~\cite{lameski2015svm}, trail $\beta_t$ for assigning proper rank is set to 15. We set the starting value of $\beta$ to 300, which the value is adjusted over iterations according to Eq.~\ref{eq:TRBS}. Also, the maximum time limit for local search is set to 5 seconds. Since a maximum run time should be given to ILP and our proposed TSA, we set the maximum time limit for both methods according to Eq.~\ref{eq:tmax} chosen to be comparable to the average running times reported for the other two heuristic approaches from the literature. It is notable that all methods (GWSA, WFC, and TSA) except the ILP are non-deterministic, so, to ensure a fair comparison, we execute each stochastic algorithm ten times per instance and reported the best, worst, and average distances, and finally, we compare them over the average hamming distance over all ten runs.  
\begin{equation}
    \label{eq:tmax}
    t_{max} (s) \leftarrow
    \begin{cases}
             30 & \text{$L < 400$;}\\
             60 & \text{$400 \leq L < 1000$;} \\
             120 & \text{$L \geq 1000$.}
    \end{cases}
\end{equation}
\subsection{Numerical Comparisons}\label{comparison}

In this section, we compare our TSA with ILP~\cite{meneses2004}, GWSA~\cite{pappalardo2014}, and WFC~\cite{xu2022} across our randomly generated and real-world datasets. Table~\ref{tab:summary} reports an overall summary results of each method. The following conclusions are drawn from there.
\begin{itemize}
    \item Among the five datasets, TSA achieved a better average solution length in four of them. For example, the average solution quality between the TSA and the second-best ILP approach on the benchmark set \emph{FLU-A-H1N1} is 819.72 vs. 874.6 in favor of our approach.  For the benchmark set \emph{McClure},  the best results are achieved by the ILP, slightly better than that of the WFC and TSA. The reason lies in the fact that these instances are small and the advanced exact approach like the Branch-and-cut incorporated in the CPLEX solver is extremely effective here. 

    \item Regarding the number of best solutions achieved, TSA attained the highest numbers again for four datasets. For \emph{Alpha-4}, \emph{Alpha-20}, and \emph{FLU-A-H1N1} the differences are huge in favor of our approach, and slightly better on \emph{TP53} from other approaches. As expected, ILP is the best one for the benchmark set \emph{McClure} (the best results achieved in 5 out of 5 cases), and the second best TSA (the best result achieved in 3 out of 5 cases). Overall, our TSA approach achieved the best results in 148 (out of 200 cases), while the second-best ILP approach did it in 49 cases, followed by the WFC with 43 cases.

    \item In summary, concerning all datasets (comprising 200 instances), GWSA, WFC, ILP, and TSA could reach 2\% (4), 21.5\% (43), 24.5\% (49), and 74\% (148) best solutions, respectively.
\end{itemize}

\begin{table}[hptb!]
\caption{Summary results over all benchmark sets.}
\label{tab:summary}
\centering
\resizebox{\columnwidth}{!}{
\begin{tabular}{llllllllllllllllllll}
\hline
\multicolumn{2}{l}{Benchmark} && \multicolumn{2}{l}{ILP} &  & \multicolumn{2}{l}{GWSA} &  & \multicolumn{2}{l}{WFC} &  & \multicolumn{2}{l}{TSA} \\ \cline{1-2} \cline{4-5} \cline{7-8} \cline{10-11} \cline{13-14} 
    Name    & \texttt{\#}   & &  $\mu (|CS|)$  &   \texttt{\#}best  & &   $\mu (|CS|)$ &  \texttt{\#}best   &  &  $\mu (|CS|)$ & \texttt{\#}best &    &   $\mu (|CS|)$ &  \texttt{\#}best       \\
    \hline
        \textit{Alpha-4}     & 88    && 353       & 27           && 315.99 &  0    &&  309.34 &  3  &&  \textbf{307.32}   &\textbf{64}  \\
        \textit{Alpha-20}    & 88    && 433.27    & 11           && 408.34 &  0    && 400.34  &  39 && \textbf{400.29}    &\textbf{64}  \\
        \textit{TP53}       & 9     && 1032      & 6            && 989.06 &  4    && 992.76  &  0  && \textbf{988.82}    &\textbf{7}   \\
        \textit{FLU-A-H1N1} & 10    && 874.6     & 0            && 942.79 &  0    && 889.76  &  0  && \textbf{819.72}    &\textbf{10}  \\
        \textit{McClure}    & 5     &&\textbf{84}& \textbf{5}   && 94.86  &  0    && 84.9    &  1  && 85.14              &3            \\
\hline  All           & 200   &&     560.72      & 49           &&  550.2      &   4   &&   535.42      &  43 &&         \textbf{520.25}           & \textbf{148}          \\
\hline
\end{tabular}}
\end{table}

The full numerical results are reported in Tables~\ref{tab:TP53}--\ref{tab:McClure} for the benchmark set \emph{TP53}, \emph{FLU-A-H1N1}, and \emph{McClue}. 
For the benchmark sets \textit{Alpha-4} and \emph{Alpha-20},  complete numerical results are reported in Appendix, see Tables~\ref{tab:Alpha4} and \ref{tab:Alpha20}, due to the sizes of the tables.

In the case of the \emph{Alpha-4} benchmark set, the following conclusions can be made.
\begin{itemize}
    \item   For small and moderate instances ($n \leq 30$), TSA achieved 4 out of 44 optimal solutions, while ILP reached 18 out of 44.   However, although TSA did not attain optimality for the remaining moderate instances, obtained solutions were very close to the quality of known optimal solutions found by the ILP ($\approx 97\%$ of the quality of the known optimal solutions).
    \item As the instance size increases ($n > 30$), the ILP approach becomes less effective, and our TSA approach outperforms all other competitor approaches.
\end{itemize}

In the case of the \emph{Alpha-20} benchmark set, the following conclusions can be made.
\begin{itemize}
    \item  While the ILP model achieved 8 optimal solutions and 11 best solutions mainly for the smallest instances ($n=10$), TSA reached 6 optimal solutions and 64 best-known solutions.
    \item For the instances with $n \in \{20, 25\}$, our TSA found a best solution in all cases. 
    \item  For the instances with $n \in \{40, 60\}$, the situation here is less clear and WFC and TSA perform similarly while significantly outperform the other two.
    \item  For the largest instances with $n \in \{80, 100\}$, TSA and WFC again perform similarly, and much better than the remaining competitors.
\end{itemize}

 Regarding the real-world \textit{TP53} dataset (Table~\ref{tab:TP53}) the following conclusions are drawn.
 \begin{itemize}
     \item  ILP, GWSA, and TSA reach six, three, and five optimal solutions, respectively. ILP has been extremely effective for $n \leq 10$. However, its performance quickly deteriorates by increasing $n$. 
     \item Overall, the TSA approach outperforms the other approaches in both the average solution quality  and the number of best solutions achieved.
 \end{itemize}

\begin{table}[h]
\caption{Comparison of all methods over \textit{TP53} benchmark}
\label{tab:TP53}
\centering
\resizebox{\columnwidth}{!}{
\begin{tabular}{llllllllllllllllllll}
\hline
\multicolumn{3}{l}{} & \multicolumn{2}{l}{ILP} &  & \multicolumn{4}{l}{GWSA} &  & \multicolumn{4}{l}{WFC} &  & \multicolumn{4}{l}{TSA} \\ \cline{4-5} \cline{7-10} \cline{12-15} \cline{17-20} 
        $|\Sigma|$   & n      &L     &solution          &t(s)     &     &best     &worst     &average     &t(s)  &  &best     &worst     &average     &t(s)  &  &best     &worst     &average     &t(s)    \\ \hline
   
      4&5    &1000      &\textbf{486*}     &1.8    &  & 486     &486     &\textbf{486*}     &49.4     &  &487     &490     &488.7     &1.8  &  &486     &486     &\textbf{486*}     &91.3    \\
      4&5    &2000      &\textbf{1003*}    &4.9    &  & 1003    &1003    &\textbf{1003*}    &181.3    &  &1005    &1011    &1008.5    &6.4  &  &1003    &1003    &\textbf{1003*}    &100.3    \\
      4&5    &3000      &\textbf{1516*}    &8.5    &  & 1516    &1516    &\textbf{1516*}    &262.4    &  &1517    &1521    &1518.7    &14.5 &  &1516    &1516    &\textbf{1516*}    &227.1    \\
      4&10   &500       &\textbf{285*}     &5.3    &  & 289     &290     &289.8   &28       &  &290     &292     &290.4     &0.5  &  &289     &289     &289     &44.8    \\
      4&10   &1000      &\textbf{580*}     &26.5   &  & 581     &582     &581.3   &54       &  &581     &583     &582.3     &2.1  &  &580     &580     &\textbf{580*}     &83.1    \\
      4&10   &2000      &\textbf{1138*}    &244.2  &  & 1139    &1140    &1139.4  &187.3    &  &1143    &1145    &1144.2    &9.1  &  &1138    &1138    &\textbf{1138*}    &164.5    \\
      4&27   &1000      &692     &253.9  &  & 648     &651     &649.4   &80       &  &652     &657     &654.6     &4.1  &  &648     &650     &\textbf{649}     &121.2    \\
      4&27   &2000      &1473    &257.6  &  & 1291    &1295    &1293.6  &238.2    &  &1297    &1301    &1298.8    &16.7 &  &1292    &1295    &\textbf{1293.3}  &141.8   \\
      4&27   &3000      &2117    &261.4  &  & 1941    &1945    &\textbf{1943.1}  &362.4    &  &1945    &1952    &1948.7    &39.6 &  &1943    &1946    &1945.1  &139.1   \\
      \hline
\end{tabular}}
\end{table}

For the second real-world dataset, \textit{FLU-A-H1N1} (Table~\ref{tab:flu}) the following conclusions are made.
\begin{itemize}
    \item ILP could not solve any of the instances to optimality. This is mainly because for a larger $n$ (in this case $n=20$) this approach is already inefficient. 
    \item  TSA is the clear winner, outperforming the other methods in all instance. 
    \item The conservation percentages of the obtained solutions range from 38 to 40 \%.  
\end{itemize}

\begin{table}[hptb!]
\caption{Comparison of all methods over \textit{FLU-A-H1N1 dataset} benchmark}
\label{tab:flu}
\resizebox{\columnwidth}{!}{
\begin{tabular}{llllllllllllllllllllll}
\hline
\multicolumn{3}{l}{} & \multicolumn{2}{l}{ILP} &  & \multicolumn{4}{l}{GWSA} &  & \multicolumn{4}{l}{WFC} &  & \multicolumn{4}{l}{TSA} &  & Conservation \\ \cline{4-5} \cline{7-10} \cline{12-15} \cline{17-20} \cline{22-22} 
  Instance    &   n    &  l    &     solution      & t(s)         &  &best     &worst     &average     &t(s)    &  &best     &worst     &average     &t(s)    &  &best     &worst     &average     &t(s)    &  & percent (\%) \\ \hline
      4-NA(6)-1000&20       &1000      &620    &123.8  &  &685   &690   &687.2  &344.3 &  &643     &652  &647.8  &3.1    &  &615  &616  &\textbf{615.7} &189.3 & &38   \\
      4-NA(6)-1100&20       &1100      &689    &123.2  &  &775   &779   &777.1  &416.8 &  &715     &721  &717.4  &4.3    &  &671  &672  &\textbf{671.4} &169.2 & &39   \\
      4-NA(6)-1200&20       &1200      &730    &123.7  &  &824   &825   &824.1  &496.3 &  &751     &755  &753.3  &4.5    &  &728  &728  &\textbf{728}   &107.1 & &39   \\
      4-NA(6)-1300&20       &1300      &865    &123.5  &  &895   &898   &896.4  &587.2 &  &835     &840  &837.4  &5.3    &  &801  &803  &\textbf{802.3} &110.7 & &38   \\
      4-NA(6)-1365&20       &1365      &944    &124.1  &  &942   &946   &944.3  &635.2 &  &868     &876  &872.1  &6.2    &  &836  &837  &\textbf{836.9} &116.9 & &38   \\
      4-HA(4)-1300&20       &1300      &810    &123.8  &  &891   &892   &891.3  &574.1 &  &858     &862  &859.7  &5.3    &  &795  &796  &\textbf{795.8} &115.7 & &38   \\
      4-HA(4)-1400&20       &1400      &891    &123.9  &  &1009  &1012  &1010.6 &662.9 &  &960     &965  &961.8  &6.1    &  &847  &847  &\textbf{847}   &119.5 & &39   \\
      4-HA(4)-1500&20       &1500      &938    &124.3  &  &1049  &1051  &1050   &764.2 &  &1000    &1007 &1004   &7.4    &  &907  &908  &\textbf{907.4} &117.6 & &39  \\
      4-HA(4)-1600&20       &1600      &1092   &124.5  &  &1132  &1138  &1134.6 &855.4 &  &1080    &1088 &1083.9 &8.4    &  &973  &975  &\textbf{973.4} &127.2 & &39  \\
      4-HA(4)-1700&20       &1700      &1167   &124.7  &  &1212  &1213  &1212.3 &959.4 &  &1158    &1163 &1160.2 &9.2    &  &1019 &1020 &\textbf{1019.3}&116.9 & &40  \\ \hline
\end{tabular}}
\end{table}

For the \textit{McClure} dataset (Table~\ref{tab:McClure}), we conclude the following.

\begin{itemize}
    \item The ILP approach solved all (5) instances optimally. 
    \item Concerning average solution quality, WFC took the second place, and TSA came at the third place. 
    \item Nevertheless, TSA achieved more best solutions than WFC. 
\end{itemize}

\begin{table}[hptb!]
\caption{Comparison of all methods over \textit{McClure} benchmark}
\label{tab:McClure}
\centering
\resizebox{\columnwidth}{!}{
\begin{tabular}{llllllllllllllllllll}
\hline
\multicolumn{1}{l}{} & \multicolumn{2}{l}{ILP} &  & \multicolumn{4}{l}{GWSA} &  & \multicolumn{4}{l}{WFC} &  & \multicolumn{4}{l}{TSA} \\ \cline{2-3} \cline{5-8} \cline{10-13} \cline{15-18} 
        Instance    &solution          &t(s)     &     &best     &worst     &average     &t(s)  &  &best     &worst     &average     &t(s)  &  &best     &worst     &average     &t(s)    \\ \hline

    McClure-586-20-6-100&	\textbf{72*}&		0.3&&	77&	79&	77.7&	5.3&&	72&	72&	\textbf{72*}&	0.04&&	72&	72&	\textbf{72*}&	16.1 \\
    McClure-586-20-10-98&	\textbf{75*}&	2.1&&	78&	79&	78.1&	5.1&&	76&	77&	76.3&	0.07&&	75&	75&	\textbf{75*}&	21.3 \\
    McClure-586-20-12-98&	\textbf{77*}&	26.1&&	80&	82&	81.3&	4.7&&	77&	79&	78&	0.07&&	77&	77&	\textbf{77*}&	25.7 \\
    McClure-582-20-6-141&	\textbf{\textemdash}&\textbf{\textemdash}	&& \textbf{\textemdash}	&\textbf{\textemdash}	&\textbf{\textemdash}	&\textbf{\textemdash}	&&	\textbf{\textemdash}&	\textbf{\textemdash}&\textbf{\textemdash}	&\textbf{\textemdash}	&&	\textbf{\textemdash}&\textbf{\textemdash}	&\textbf{\textemdash}	&	\textbf{\textemdash} \\
    McClure-582-20-10-141&	\textbf{97*}&	1.3&&	120&	125&	122.4&	7.2&&	98&	100&	99.3&	0.09&&	101&	101&	101&	18.1 \\
    McClure-582-20-12-141&	\textbf{97*}&	3.2&&	114&	116&	114.8&	10.8&&	98&	100&	98.9&	0.1&&	100&	101&	100.7&	18.6 \\
    \hline    
\end{tabular}}
\end{table}

\subsection{Time Comparison}\label{time comparison}
\begin{figure}[H]
     \centering
     \begin{subfigure}[b]{0.45\textwidth}
         \centering
         \includegraphics[width=\textwidth]{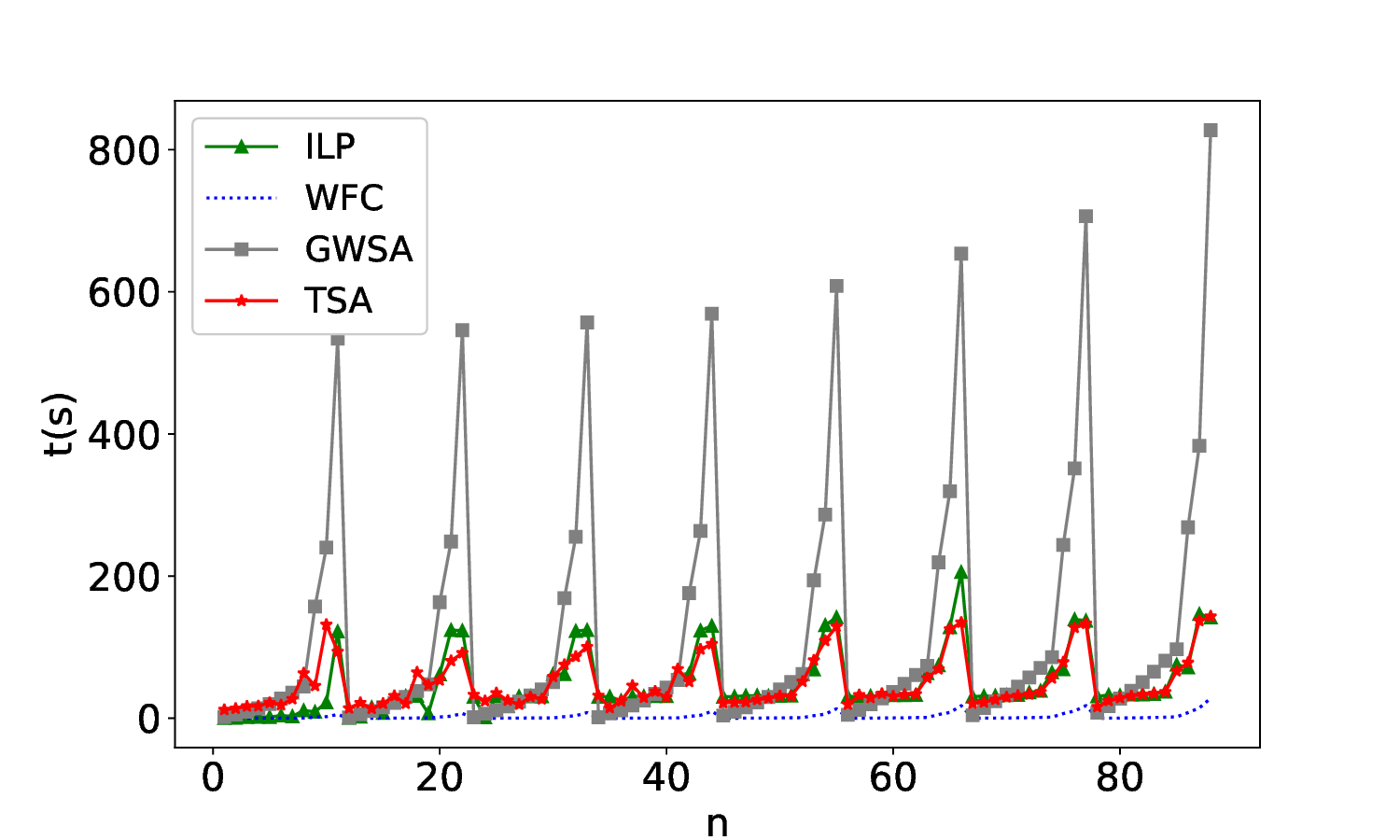}
         \caption{Average runtimes over \textit{Alpha-4}.}
         \label{fig:time:Alpha-4}
     \end{subfigure}
     \hfill
     \begin{subfigure}[b]{0.45\textwidth}
         \centering
         \includegraphics[width=\textwidth]{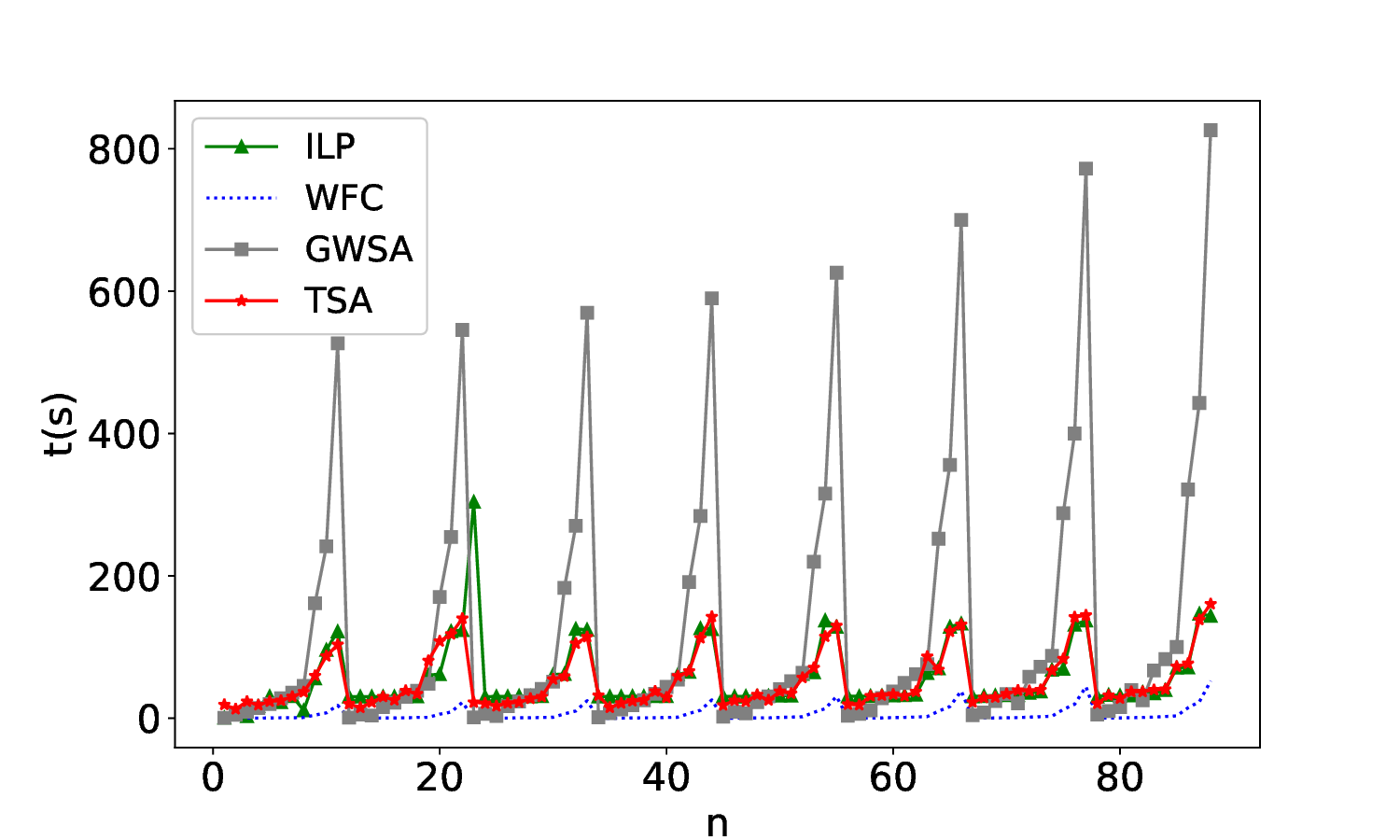}
         \caption{Average runtimes over \textit{Alpha-20}.}
         \label{fig:time:Alpha-20}
     \end{subfigure}
     \hfill
    
     \begin{subfigure}[]{0.45\textwidth}
         \centering
         \includegraphics[width=\textwidth]{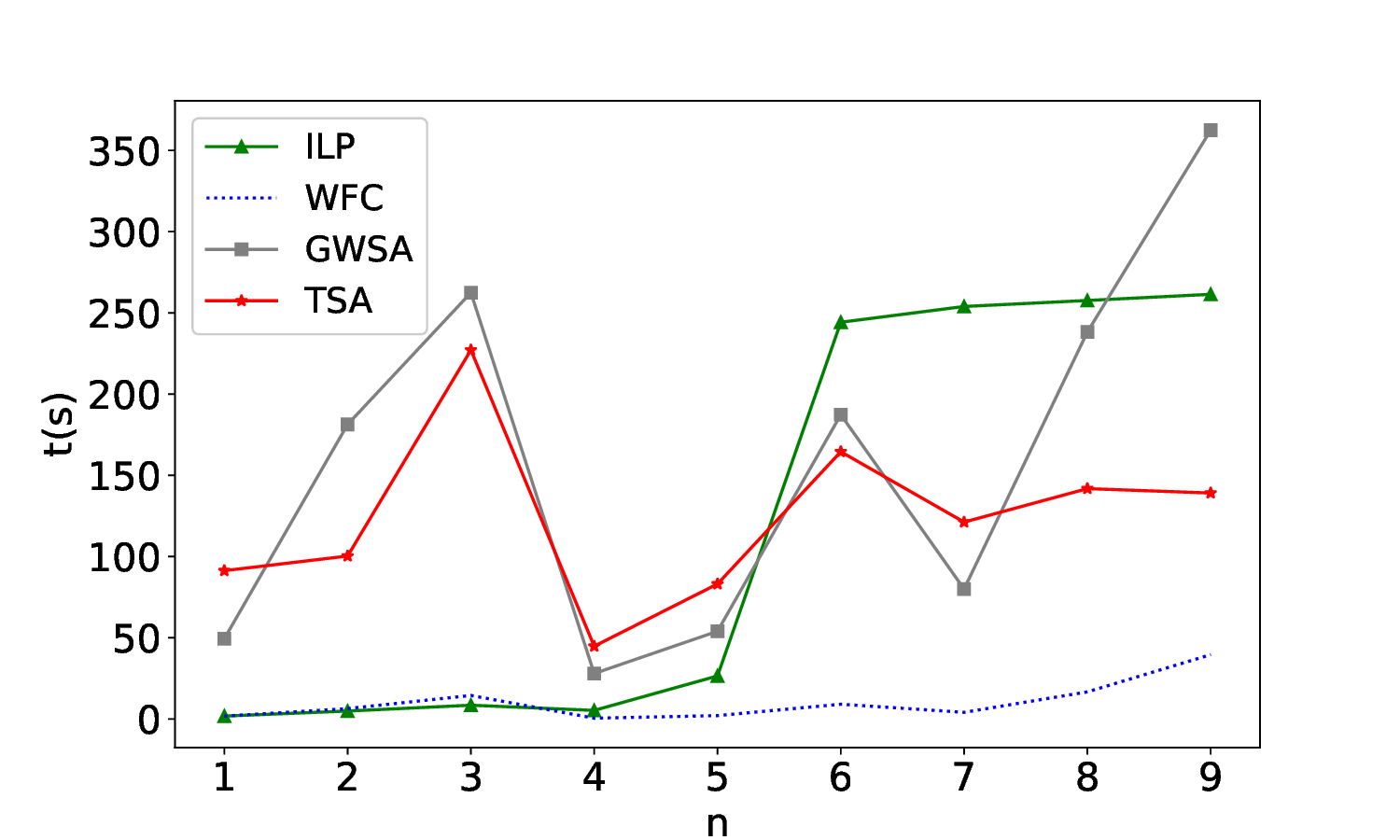}
         \caption{Average runtimes over \textit{TP53}. }
         \label{fig:time:TP53}
     \end{subfigure}
     \hfill
     \begin{subfigure}[]{0.45\textwidth}
         \centering
         \includegraphics[width=\textwidth]{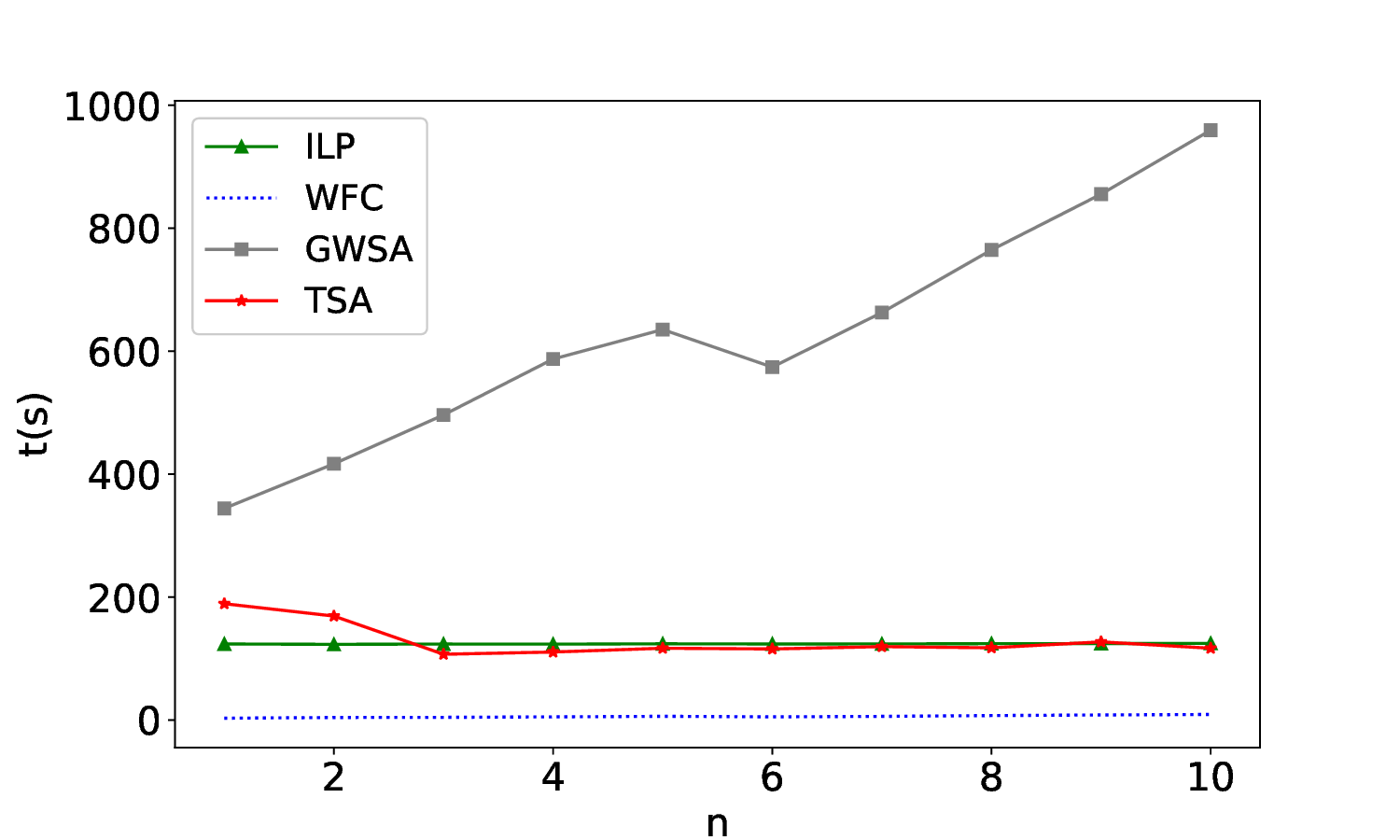}
         \caption{Average runtimes over \textit{FLU-A-H1N1}}
         \label{fig:time:FLU-A-H1N1}
     \end{subfigure}
     
     \begin{subfigure}[]{0.45\textwidth}
         \centering
         \includegraphics[width=\textwidth]{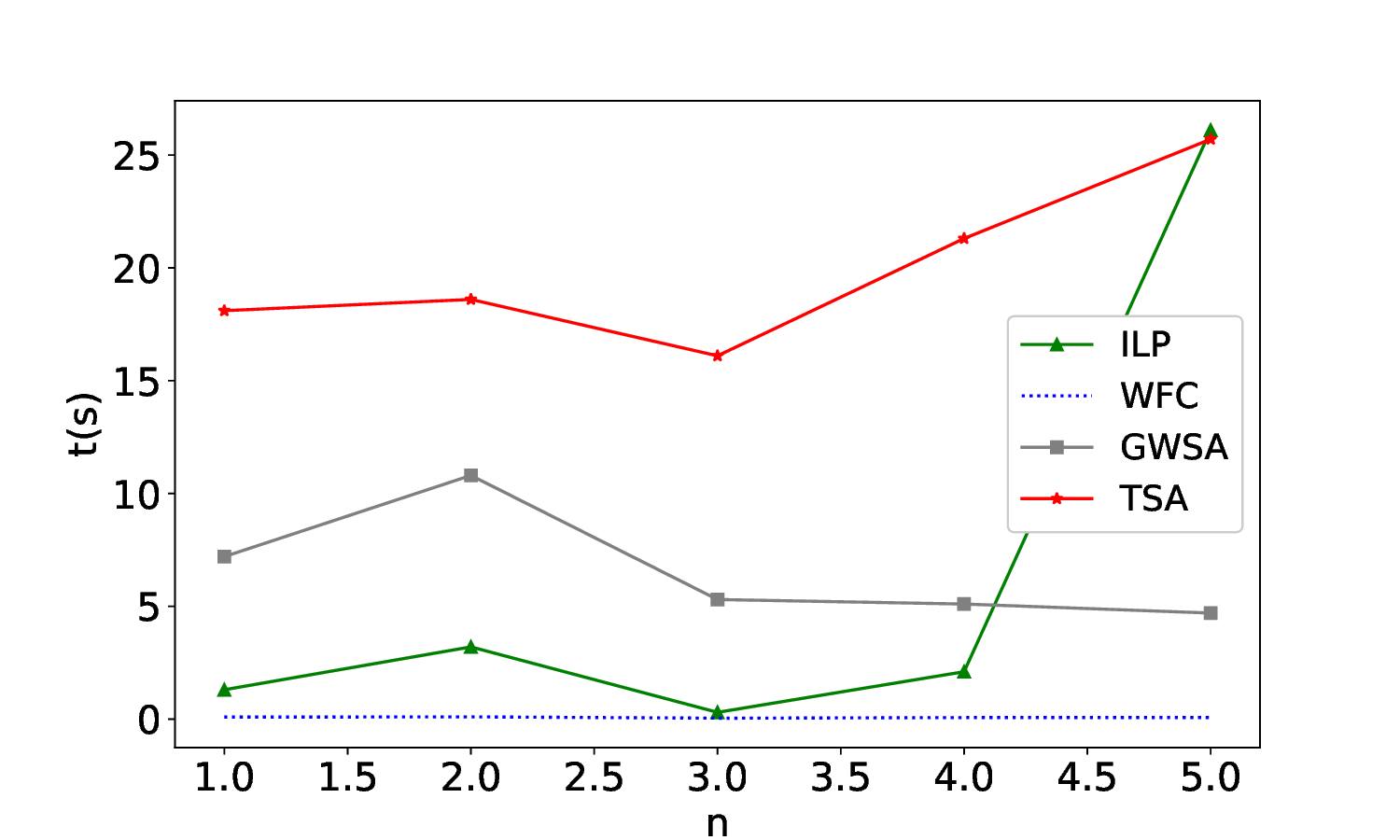}
         \caption{Average runtimes over \textit{McClure}}
         \label{fig:time:McClure}
     \end{subfigure}
        \caption{Time plots of all approaches over all five benchmark datasets}
        \label{fig:time}
\end{figure}

In this section, we compare the algorithms from a running time perspective. Figure~\ref{fig:time} illustrates the run times of all methods across five benchmarks. In each plot, the $y$-axis  represents the running time (in seconds), while the $x$-axis indicates the instance number (a unique instance ID) indicating an instance for which the (average) running time is obtained by executing respective algorithm.

The WFC algorithm exhibits the lowest running time, however it comes with poor performance. Due to the absence of tunable parameters, we consider this method to have reached its limitations. For the \textit{Alpha-4}, \textit{Alpha-20}, and \textit{FLU-A-H1N1} datasets, ILP and TSA algorithms demonstrate similar running times. However, for the \textit{TP53} and \textit{McClure} datasets, TSA outperforms ILP on medimum to large-sized instances, while ILP is faster for smaller instances where instances are solved optimally. When compared to the GWSA method, the TSA generally has a shorter running time, with the notable exception of the \textit{McClure} dataset.

\subsection{Statistical Analysis}\label{statistical}

In this section, we compare methods from a statistical perspective. For this analysis, we utilize the Friedman test~\cite{friedman1940} at a 5\% significance level for the results of all four competitors over two groups of instances: the first group includes randomly generated instances from \emph{Alpha}-4 and \emph{Alpha}-20 (176 instances), whereas the second group includes real-world instances from the remaining three benchmark sets (24 instances in overall).

\begin{figure}[H]
     \centering
     \begin{subfigure}[]{0.85\textwidth}
         \centering
         \includegraphics[width=\textwidth]{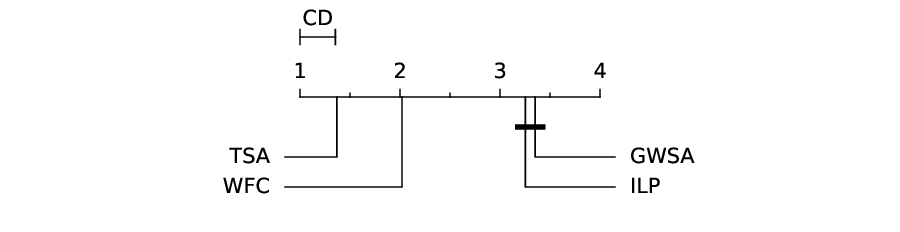}
         \caption{All methods over 176 random instances}
         \label{fig:stat:alpha-4_IL}
     \end{subfigure}
     \begin{subfigure}[]{0.85\textwidth}
         \centering
         \includegraphics[width=\textwidth]{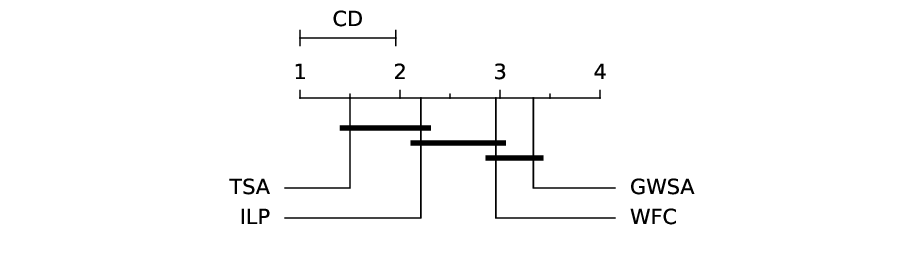}
         \caption{All methods over 24 real-world instances}
         \label{fig:stat:alpha-4_WFC}
     \end{subfigure}
     \caption{Critical difference plot concerning the quality of solutions over all benchmarks}
    \label{fig:overall_stat}
\end{figure}

The results of the Friedman test performed for both groups separately indicate that we reject the null hypothesis ($H_0$), suggesting that the methods do not perform equally. Subsequently, the pairwise comparison of methods is illustrated in Figure~\ref{fig:overall_stat} using CD plots by utilizing Nemenyi post-hoc test~\cite{pohlert2014pairwise}. 
In detail, for each pair of algorithms, a critical difference is calculated at the significance level of 5\%. Each algorithm is positioned at $x$-axis w.r.t. its average ranking. If there is a horizontal bar joining two algorithms, it means that they perform statistically equally.

The following conclusions are drawn from the generated CD plots. 

\begin{itemize}
    \item Concerning random instances, the best average ranking is obtained by our TSA. The second-best average ranking is achieved by WFC, followed by ILP and GWSA. 
    Hereby, the TSA approach performs significantly better than the competitor WFC. Both of these approaches are statistically significantly better than ILP and GWSA. 
    
    \item Concerning real-world instances,  TSA again achieved the best average ranking (1.4), followed by the ILP approach. However, here the performance of both approaches does not differentiate significantly from each other.  On the other hand,  TSA outperforms the GSWA and WFC statistically significantly. 
    However, no statistical difference is presented between the results of ILP and WFC.
\end{itemize}

\section{Conclusion}\label{conc}

In this paper, we introduced the Three-Stage Algorithm for solving the Closest String problem. Our proposed method comprises three core stages. First, we introduce a novel pruning to reduce the search space effectively. In particular,  at each level of the search, a predetermined subset of promising members of the alphabet instead of a whole set is employed for a set of allowed transitions. This pruning method positively affects the quality of solutions while reducing the overall running time of the algorithm. Second, a time-restricted beam search to explore the search space for obtaining complete solutions is utilized. The search process of the beam search is guided by the newly designed expected distance heuristic function. Third, a new local search algorithm is proposed. It enables a complete solution to be improved by repetitively substituting critical characters in the incumbent solution. In addition, we created four datasets: two real-world and two uniformly at random distributed to test the robustness of the methods on both, real and simulated sequences. 

Our extensive experimental results show that the proposed method can significantly improve over the best solutions from the literature thus yielding new state-of-the-art method. In some details, for both, random and real-world instances, the results of our method achieved the best average ranking when compared to other competitor approaches. In the case of random instances, the difference between our approach and the second-best  WFC approach is statistically significant in favor of the former approach. In the case of real-world instances, our approach and the second-best, an ILP approach, perform statistically equally good, and significantly better than the other two approaches. 

Regarding future work, one could engage the Monte Carlo Tree Search in the search process to leverage a predictive exploration in the state space of the CS problem and to design possibly more accurate heuristic guidance. Another way could be designing a complementary approach to the classical optimization approaches. This includes designing a learning mechanism that can be developed by separately training a model of machine learning, such as a kind of neural network, to support better decision-making in the beam search, see e.g.~\cite{reixach2024neural}. In order to do so,  one has to determine a set of crucial input features  extracted from instances that affect the final score; e.g. maximum frequency of characters at each level, standard deviation/average of maximum frequencies per each level, the variance of the solution w.r.t. related values of the hamming distance to the input strings, etc.

\section*{Acknowledgements} Marko Djukanović acknowledges financial support of the  Ministry of Science and Technology Development and Higher Education of the Republic of Srpska with the project entitled ``Development of artificial intelligence models and algorithms for solving difficult combinatorial optimization problems'' 
under no. 1259086. 


\bibliography{\jobname.bib}

\newpage
\appendix

\section{Numerical Results for Benchmark Set \emph{Alpha-4}}

{\small\tabcolsep=2pt
\small
\begin{longtable}{llllllllllllllllllll}
\caption{Comparison of all methods over \textit{Alpha-4} benchmark}
\label{tab:Alpha4} \\
\hline
\multicolumn{3}{l}{} & \multicolumn{2}{l}{ILP} &  & \multicolumn{4}{l}{GWSA} &  & \multicolumn{4}{l}{WFC} &  & \multicolumn{4}{l}{TSA} \\ \cline{4-5} \cline{7-10} \cline{12-15} \cline{17-20} 
        $|\Sigma|$   & n      &l      &solution          &t(s)     &     &best     &worst     &average     &t(s)  &  &best     &worst     &average     &t(s)  &  &best     &worst     &average     &t(s)    \\ \hline
\endfirsthead
    
\caption{Comparison of all methods over \textit{Alpha-4} benchmark} \\
\hline
\multicolumn{3}{l}{} & \multicolumn{2}{l}{ILP} &  & \multicolumn{4}{l}{GWSA} &  & \multicolumn{4}{l}{WFC} &  & \multicolumn{4}{l}{TSA} \\ \cline{4-5} \cline{7-10} \cline{12-15} \cline{17-20} 
        $|\Sigma|$   & n      &l      &solution          &t(s)     &     &best     &worst     &average     &t(s)  &  &best     &worst     &average     &t(s)  &  &best     &worst     &average     &t(s)    \\ \hline
\endhead
\endfoot    
\hline    
\endlastfoot 

    4 & 10 & 50 & \textbf{31*} & 0.3 & & 34 & 35 & 34.1 & 1.7 && 32 & 32 & 32 & 0.005 && 31 & 31 & \textbf{31*} & 11.9  \\
    4 & 10 & 100 & \textbf{58*} & 0.5& & 61 & 63 & 62.4 & 5 && 59 & 60 & 59.7 & 0.02 && 59 & 59 & 59 & 13.6  \\
    4 & 10 & 150 & \textbf{88*} & 1.6 && 92 & 94 & 93 & 9.1 && 89 & 91 & 89.8 & 0.04 && 89 & 89 & 89 & 16.8  \\
    4 & 10 & 200 & \textbf{116*} & 1.8 && 122 & 125 & 123.4 & 14 && 117 & 119 & 118.1 & 0.07 && 118 & 118 & 118 & 17.1  \\
    4 & 10 & 250 & \textbf{144*} & 1.7 && 150 & 153 & 151.5 & 19.8 && 145 & 148 & 146.5 & 0.1 && 144 & 145 & 144.4 & 21.7  \\
    4 & 10 & 300 & \textbf{175*} & 3.7 && 186 & 189 & 187.5 & 27.6 && 178 & 178 & 178 & 0.2 && 176 & 176 & 176 & 18.5 \\
    4 & 10 & 350 & \textbf{202*} & 2.6 && 211 & 215 & 212.7 & 36 && 203 & 205 & 204.1 & 0.2 && 202 & 202 & \textbf{202*} & 27.1  \\
    4 & 10 & 400 & \textbf{231*} & 10.9 && 238 & 240 & 238.8 & 44.7 && 232 & 233 & 232.7 & 0.3 && 232 & 233 & 232.3 & 63.3  \\
    4 & 10 & 800 & \textbf{467*} & 9 && 484 & 488 & 486.4 & 157.1 && 469 & 472 & 470.6 & 1.4 && 469 & 469 & 469 & 45.6  \\
    4 & 10 & 1000 & \textbf{579*} & 22.2 && 588 & 593 & 589.9 & 240.1 && 580 & 582 & 581.1 & 2.2 && 579 & 579 & \textbf{579*} & 131.8  \\
    4 & 10 & 1500 & 874 & 122 && 883 & 885 & 884.4 & 534 && 872 & 874 & 872.7 & 5.1 && 871 & 871 & \textbf{871*} & 93.7  \\
    4 & 15 & 50 & \textbf{32*} & 0.7 && 35 & 37 & 35.9 & 1 && 33 & 34 & 33.4 & 0.007 && 33 & 33 & 33 & 14  \\
    4 & 15 & 100 & \textbf{63*} & 2.2 && 69 & 71 & 70.5 & 5.4 && 65 & 66 & 65.2 & 0.02 && 64 & 64 & 64 & 21.7  \\
    4 & 15 & 150 & \textbf{93*} & 15.2 && 98 & 100 & 99 & 10.1 && 95 & 96 & 95.7 & 0.06 && 94 & 94 & 94 & 13.6  \\
    4 & 15 & 200 & \textbf{124*} & 7.5 && 126 & 127 & 126.7 & 15.3 && 126 & 127 & 126.5 & 0.1 && 126 & 126 & 126 & 20.6 \\
    4 & 15 & 250 & \textbf{154*} & 28.6 && 161 & 163 & 162.5 & 21.5 && 156 & 158 & 156.8 & 0.1 && 156 & 156 & 156 & 31.5  \\
    4 & 15 & 300 & \textbf{186*} & 31.2 && 194 & 198 & 195.9 & 29.9 && 187 & 190 & 188.6 & 0.2 && 188 & 188 & 188 & 20.7  \\
    4 & 15 & 350 & \textbf{215} & 31.3 && 227 & 230 & 229.2 & 38.3 && 217 & 221 & 218.8 & 0.3 && 218 & 218 & 218 & 64.6  \\
    4 & 15 & 400 & \textbf{246} & 6.8 && 255 & 257 & 256.1 & 47.6 && 248 & 249 & 248.1 & 0.4 && 246 & 246 & \textbf{246} & 46.8  \\
    4 & 15 & 800 & \textbf{493} & 61.6 && 505 & 508 & 506.1 & 163.2 && 495 & 496 & 495.6 & 1.7 && 494 & 494 & 494 & 53.4  \\
    4 & 15 & 1000 & 1001 & 124 && 632 & 637 & 633.8 & 248.6 && 619 & 621 & 620.3 & 2.8 && 618 & 618 & \textbf{618} & 80.7  \\
    4 & 15 & 1500 & 1461 & 123.2 && 940 & 945 & 942.3 & 546 && 921 & 923 & 921.8 & 6.7 && 919 & 919 & \textbf{919} & 91.7  \\
    4 & 20 & 50 & \textbf{34} & 30.1 && 39 & 41 & 40.4 & 1.3 && 35 & 36 & 35.5 & 0.009 && 34 & 34 & \textbf{34} & 33.2 \\
    4 & 20 & 100 & \textbf{65*} & 1.1 && 67 & 69 & 68.2 & 6.3 && 66 & 67 & 66.9 & 0.03 && 66 & 66 & 66 & 24.3  \\
    4 & 20 & 150 & \textbf{96} & 30.8 && 100 & 105 & 102.8 & 11.1 && 98 & 99 & 98.8 & 0.08 && 98 & 98 & 98 & 35.4  \\
    4 & 20 & 200 & \textbf{128*} & 17 && 135 & 136 & 135.5 & 16.7 && 130 & 132 & 131 & 0.1 && 130 & 130 & 130 & 25.1 \\
    4 & 20 & 250 & \textbf{159} & 30.6 && 167 & 169 & 168.1 & 23.3 && 160 & 163 & 161.4 & 0.2 && 161 & 161 & 161 & 19.4  \\
    4 & 20 & 300 & 214 & 30.8 && 199 & 203 & 200.7 & 31.9 && 195 & 197 & 195.6 & 0.3 && 193 & 193 & \textbf{193} & 30.1  \\
    4 & 20 & 350 & 245 & 30.9 && 231 & 234 & 232.4 & 40.6 && 225 & 227 & 225.6 & 0.3 && 223 & 224 & \textbf{223.5} & 26.9  \\
    4 & 20 & 400 & 281 & 62.1 && 259 & 261 & 259.7 & 50.7 && 255 & 257 & 255.7 & 0.4 && 254 & 254 & \textbf{254} & 58.4  \\
    4 & 20 & 800 & 569 & 62.1 && 523 & 528 & 525 & 169 && 513 & 515 & 514.3 & 2.2 && 513 & 515 & \textbf{513.6} & 75.4  \\
    4 & 20 & 1000 & 704 & 122.7 && 654 & 662 & 656.6 & 255.3 && 638 & 640 & 639.3 & 3.5 && 638 & 638 & \textbf{638} & 86.7  \\
    4 & 20 & 1500 & 1391 & 124 && 973 & 976 & 974.9 & 556.9 && 955 & 957 & 956.2 & 7.9 && 953 & 953 & \textbf{953} & 100.5  \\
    4 & 25 & 50 & \textbf{35} & 30.2 && 39 & 41 & 40.2 & 1.3 && 36 & 36 & 36 & 0.009 && 35 & 35 & \textbf{35} & 31.8  \\
    4 & 25 & 100 & \textbf{68} & 30.4 && 71 & 74 & 72.1 & 6.7 && 69 & 70 & 69.5 & 0.03 && 69 & 69 & 69 & 14.6  \\
    4 & 25 & 150 & \textbf{98*} & 24.6 && 106 & 109 & 106.8 & 11 && 101 & 102 & 101.9 & 0.07 && 100 & 100 & 100 & 24.1  \\
    4 & 25 & 200 & 136 & 30.6 && 133 & 135 & 134.2 & 18.1 && 131 & 134 & 132.3 & 0.1 && 131 & 131 & \textbf{131} & 45.9  \\
    4 & 25 & 250 & 173 & 30.8 && 171 & 174 & 172.3 & 25 && 166 & 168 & 167.2 & 0.1 && 166 & 166 & \textbf{166} & 29.1  \\
    4 & 25 & 300 & 213 & 31 && 208 & 210 & 209.1 & 34.1 && 198 & 200 & 198.8 & 0.3 && 198 & 199 & \textbf{198.7} & 37.7  \\
    4 & 25 & 350 & 246 & 31.1 && 235 & 239 & 237.6 & 43.1 && 229 & 231 & 230.4 & 0.4 && 227 & 227 & \textbf{227} & 29.4  \\
    4 & 25 & 400 & 288 & 61.2 && 266 & 269 & 266.9 & 53.8 && 262 & 264 & 262.6 & 0.5 && 261 & 261 & \textbf{261} & 69.3  \\
    4 & 25 & 800 & 600 & 62.6 && 536 & 538 & 536.6 & 176 && 521 & 525 & 522.7 & 2.5 && 520 & 521 & \textbf{520.7} & 51.4 \\
    4 & 25 & 1000 & 678 & 123.3 && 672 & 677 & 674.1 & 263.4 && 649 & 651 & \textbf{650.2} & 4.1 && 650 & 651 & 650.3 & 97  \\
    4 & 25 & 1500 & 1501 & 130 && 996 & 1000 & 997.9 & 569.2 && 982 & 984 & \textbf{982.4} & 9.7 && 981 & 984 & 983 & 104.5  \\  

    4 & 40 & 50 & 38 & 30.2 && 38 & 39 & 38.6 & 3.7 && 37 & 38 & 37.3 & 0.01 && 36 & 36 & \textbf{36} & 21.4  \\
    4 & 40 & 100 & 77 & 30.5 && 74 & 76 & 75.1 & 9.1 && 72 & 73 & 72.6 & 0.05 && 71 & 71 & \textbf{71} & 22.3  \\
    4 & 40 & 150 & 115 & 31.6 && 109 & 111 & 110.3 & 15.2 && 105 & 107 & 106.2 & 0.1 && 104 & 104 & \textbf{104} & 22.9  \\
    4 & 40 & 200 & 146 & 31.8 && 142 & 144 & 142.7 & 22.2 && 140 & 141 & 140.1 & 0.2 && 138 & 138 & \textbf{138} & 26.1  \\
    4 & 40 & 250 & 185 & 31.3 && 177 & 179 & 178.3 & 29.9 && 175 & 176 & 175.5 & 0.3 && 175 & 175 & \textbf{175} & 28.1  \\
    4 & 40 & 300 & 222 & 31.5 && 212 & 215 & 213.3 & 40.4 && 207 & 209 & 207.9 & 0.4 && 204 & 204 & \textbf{204} & 31.3  \\
    4 & 40 & 350 & 257 & 31.8 && 247 & 251 & 248.8 & 50.9 && 242 & 244 & 243 & 0.6 && 241 & 241 & \textbf{241} & 31.1  \\
    4 & 40 & 400 & 282 & 62.1 && 278 & 282 & 280 & 62.2 && 275 & 276 & 275.3 & 0.8 && 273 & 273 & \textbf{273} & 51.6  \\
    4 & 40 & 800 & 801 & 68.4 && 550 & 554 & 551.5 & 194 && 545 & 547 & 545.9 & 3.6 && 540 & 541 & \textbf{540.1} & 81.6  \\
    4 & 40 & 1000 & 703 & 130.9 && 691 & 694 & 692.5 & 286.5 && 678 & 681 & 678.8 & 5.7 && 676 & 677 & \textbf{676.9} & 109  \\
    4 & 40 & 1500 & 1501 & 141.9 && 1027 & 1031 & 1028.4 & 608.3 && 1014 & 1017 & 1015.2 & 13.3 && 1014 & 1015 & \textbf{1014.5} & 129.2  \\
    4 & 60 & 50 & 41 & 30.4 && 39 & 42 & 40.2 & 4.7 && 38 & 39 & 38.9 & 0.01 && 37 & 37 & \textbf{37} & 19.4  \\
    4 & 60 & 100 & 79 & 30.7 && 74 & 78 & 76.3 & 11.8 && 74 & 74 & 74 & 0.07 && 72 & 72 & \textbf{72} & 32.6  \\
    4 & 60 & 150 & 117 & 31.2 && 115 & 119 & 116.2 & 19.5 && 109 & 110 & 109.8 & 0.1 && 109 & 109 & \textbf{109} & 28.7  \\
    4 & 60 & 200 & 149 & 31.5 && 147 & 148 & 147.6 & 27.8 && 143 & 145 & 143.8 & 0.3 && 142 & 142 & \textbf{142} & 34.6  \\
    4 & 60 & 250 & 251 & 32.3 && 184 & 188 & 185.8 & 37 && 179 & 180 & 179.3 & 0.4 && 177 & 177 & \textbf{177} & 30.9  \\
    4 & 60 & 300 & 230 & 32.5 && 215 & 217 & 216.3 & 48.4 && 210 & 212 & 211.3 & 0.6 && 209 & 210 & \textbf{209.4} & 33.1  \\
    4 & 60 & 350 & 260 & 32.8 && 251 & 252 & 251.1 & 60.7 && 247 & 249 & 247.8 & 0.9 && 245 & 245 & \textbf{245} & 34  \\
    4 & 60 & 400 & 293 & 63.3 && 288 & 292 & 289.5 & 73.7 && 284 & 288 & 285.5 & 1.1 && 281 & 281 & \textbf{281} & 56.7  \\
    4 & 60 & 800 & 581 & 74.4 && 571 & 577 & 573.8 & 219.3 && 558 & 561 & 559.7 & 5.2 && 556 & 556 & \textbf{556} & 69.1  \\
    4 & 60 & 1000 & 718 & 128.1 && 706 & 708 & 707.3 & 319.2 && 695 & 697 & 696.2 & 8.1 && 692 & 693 & \textbf{692.7} & 125.4  \\
    4 & 60 & 1500 & 1501 & 205.6 && 1051 & 1056 & 1053.1 & 653.7 && 1040 & 1043 & 1041.6 & 17.6 && 1035 & 1036 & \textbf{1035.9} & 134.7  \\
    4 & 80 & 50 & 41 & 30.5 && 40 & 42 & 41.1 & 4.1 && 39 & 40 & 39.7 & 0.02 && 37 & 37 & \textbf{37} & 20.8  \\
    4 & 80 & 100 & 80 & 31.1 && 78 & 80 & 79.1 & 14.5 && 76 & 77 & 76.4 & 0.09 && 74 & 74 & \textbf{74} & 22.4  \\
    4 & 80 & 150 & 113 & 31.5 && 113 & 115 & 113.9 & 23.7 && 110 & 112 & 111.2 & 0.2 && 109 & 109 & \textbf{109} & 26.3  \\
    4 & 80 & 200 & 152 & 32.4 && 151 & 152 & 151.6 & 33.3 && 147 & 148 & 147.5 & 0.3 && 144 & 145 & \textbf{144.6} & 29.7  \\
    4 & 80 & 250 & 192 & 32.6 && 185 & 187 & 186.5 & 44.6 && 183 & 184 & 183.1 & 0.6 && 180 & 181 & \textbf{180.9} & 31.8  \\
    4 & 80 & 300 & 224 & 36.2 && 219 & 221 & 220.2 & 57.2 && 217 & 219 & 217.7 & 0.8 && 215 & 215 & \textbf{215} & 33.7  \\
    4 & 80 & 350 & 258 & 38.7 && 254 & 256 & 255.5 & 70.7 && 250 & 252 & 251.3 & 1.2 && 249 & 249 & \textbf{249} & 37  \\
    4 & 80 & 400 & 299 & 64.5 && 290 & 294 & 291.4 & 85.8 && 287 & 289 & 287.9 & 1.5 && 284 & 285 & \textbf{284.6} & 56.3  \\
    4 & 80 & 800 & 594 & 68.8 && 580 & 585 & 583.2 & 243.9 && 570 & 573 & 571.4 & 6.5 && 566 & 568 & \textbf{567.4} & 78.9  \\
    4 & 80 & 1000 & 725 & 139 && 713 & 715 & 714.2 & 351.4 && 705 & 708 & 706.4 & 10.2 && 702 & 703 & \textbf{702.4} & 127.5  \\
    4 & 80 & 1500 & 1092 & 137.3 && 1062 & 1065 & 1063.6 & 706.5 && 1055 & 1057 & \textbf{1056} & 17.7 && 1056 & 1057 & 1056.7 & 133.2  \\
    4 & 100 & 50 & 42 & 30.7 && 41 & 42 & 41.5 & 7.8 && 40 & 41 & 40.7 & 0.02 && 39 & 39 & \textbf{39} & 16.1  \\
    4 & 100 & 100 & 82 & 32.6 && 82 & 86 & 83.4 & 17.1 && 77 & 79 & 77.7 & 0.1 && 75 & 76 & \textbf{75.3} & 24.5  \\
    4 & 100 & 150 & 119 & 32 && 116 & 119 & 117.2 & 27.6 && 114 & 116 & 114.5 & 0.2 && 111 & 112 & \textbf{111.2} & 28.5  \\
    4 & 100 & 200 & 156 & 33.5 && 156 & 159 & 157.6 & 38.6 && 149 & 150 & 149.7 & 0.4 && 148 & 148 & \textbf{148} & 31.4  \\
    4 & 100 & 250 & 191 & 33.5 && 190 & 192 & 191.1 & 50.8 && 185 & 187 & 186.2 & 0.6 && 182 & 183 & \textbf{182.8} & 33.3  \\
    4 & 100 & 300 & 235 & 34 && 224 & 226 & 225.4 & 65.5 && 220 & 222 & 221 & 1.1 && 218 & 218 & \textbf{218} & 34.7  \\
    4 & 100 & 350 & 265 & 37.2 && 259 & 262 & 260 & 80.8 && 256 & 258 & 256.6 & 1.4 && 252 & 253 & \textbf{252.2} & 36.9  \\
    4 & 100 & 400 & 304 & 75.3 && 297 & 300 & 298.2 & 97.1 && 292 & 294 & 292.7 & 1.9 && 289 & 289 & \textbf{289} & 66.3  \\
    4 & 100 & 800 & 585 & 71.4 && 581 & 583 & 582.2 & 268.6 && 574 & 576 & 575.1 & 7.8 && 569 & 572 & \textbf{570.2} & 78.2  \\
    4 & 100 & 1000 & 727 & 146.5 && 719 & 722 & 720.8 & 383.3 && 712 & 715 & 713.7 & 14.4 && 710 & 711 & \textbf{710.6} & 137.1  \\
    4 & 100 & 1500 & 1097 & 141.8 && 1077 & 1081 & 1079.2 & 827.4 && 1069 & 1071 & 1070 & 28.1 && 1063 & 1066 & \textbf{1064.3} & 143.2
\end{longtable}}

\clearpage
\section{Numerical Results for Benchmark Set \emph{Alpha-20}}

{\footnotesize\tabcolsep=2pt
\small
\begin{longtable}{llllllllllllllllllll}
\caption{Comparison of all methods over \textit{Alpha-20} benchmark}
\label{tab:Alpha20} \\
\hline
\multicolumn{3}{l}{} & \multicolumn{2}{l}{ILP} &  & \multicolumn{4}{l}{GWSA} &  & \multicolumn{4}{l}{WFC} &  & \multicolumn{4}{l}{TSA} \\ \cline{4-5} \cline{7-10} \cline{12-15} \cline{17-20} 
        $|\Sigma|$   & n      &l      &solution          &t(s)     &     &best     &worst     &average     &t(s)  &  &best     &worst     &average     &t(s)  &  &best     &worst     &average     &t(s)    \\ \hline
\endfirsthead
    
\caption{Comparison of all methods over \textit{Alpha-20} benchmark} \\
\hline
\multicolumn{3}{l}{} & \multicolumn{2}{l}{ILP} &  & \multicolumn{4}{l}{GWSA} &  & \multicolumn{4}{l}{WFC} &  & \multicolumn{4}{l}{TSA} \\ \cline{4-5} \cline{7-10} \cline{12-15} \cline{17-20} 
        $|\Sigma|$   & n      &l      &solution          &t(s)     &     &best     &worst     &average     &t(s)  &  &best     &worst     &average     &t(s)  &  &best     &worst     &average     &t(s)    \\ \hline
\endhead
\endfoot    
\hline    
\endlastfoot

    20 & 10 & 50 & \textbf{39*} & 0.8 &  & 45 & 45 & 45 & 0.6 &  & 40 & 40 & 40 & 0.02 &  & 40 & 40 & 40 & 19  \\
    20 & 10 & 100 & \textbf{79*} & 7.3 &  & 84 & 85 & 84.3 & 4.9 &  & 79 & 80 & 79.2 & 0.09 &  & 79 & 79 & \textbf{79*} & 12.9  \\
    20 & 10 & 150 & \textbf{117*} & 2.8 &  & 122 & 123 & 122.9 & 9 &  & 118 & 118 & 118 & 0.1 &  & 117 & 117 & \textbf{117*} & 23.5  \\
    20 & 10 & 200 & \textbf{156*} & 16.8 &  & 160 & 162 & 160.9 & 14.2 &  & 156 & 157 & 156.5 & 0.2 &  & 156 & 156 & \textbf{156*} & 18.6  \\
    20 & 10 & 250 & \textbf{196} & 30.7 &  & 206 & 208 & 206.9 & 19.9 &  & 197 & 197 & 197 & 0.3 &  & 196 & 196 & \textbf{196} & 23.6  \\
    20 & 10 & 300 & \textbf{235*} & 22.6 &  & 243 & 246 & 244.2 & 28 &  & 236 & 237 & 236.1 & 0.6 &  & 236 & 236 & \textbf{236} & 24.8  \\
    20 & 10 & 350 & \textbf{277} & 31 &  & 284 & 286 & 284.8 & 36 &  & 277 & 278 & 277.8 & 0.6 &  & 277 & 277 & \textbf{277} & 30.4  \\
    20 & 10 & 400 & \textbf{313*} & 10.9 &  & 317 & 320 & 318.5 & 45.3 &  & 313 & 314 & 313.4 & 1.1 &  & 313 & 313 & \textbf{313*} & 36.8  \\
    20 & 10 & 800 & \textbf{625*} & 55.9 &  & 637 & 639 & 638.1 & 161.4 &  & 626 & 627 & 626.3 & 4.4 &  & 625 & 625 & \textbf{625*} & 59.7  \\
    20 & 10 & 1000 & \textbf{781*} & 96 &  & 793 & 799 & 796.3 & 241.4 &  & 781 & 782 & 781.5 & 7.5 &  & 781 & 781 & \textbf{781*} & 87.2  \\
    20 & 10 & 1500 & 1185 & 122 &  & 1191 & 1197 & 1193.5 & 526.7 &  & 1176 & 1177 & 1176.3 & 18.1 &  & 1176 & 1176 & \textbf{1176} & 103.4  \\
    20 & 15 & 50 & \textbf{42} & 30.3 &  & 46 & 46 & 46 & 0.8 &  & 42 & 43 & 42.6 & 0.01 &  & 42 & 42 & \textbf{42} & 20.1  \\
    20 & 15 & 100 & 85 & 30.4 &  & 86 & 89 & 88.3 & 5.3 &  & 83 & 83 & 83 & 0.06 &  & 82 & 82 & \textbf{82} & 14.9  \\
    20 & 15 & 150 & 125 & 30.4 &  & 129 & 132 & 130.4 & 3.8 &  & 123 & 124 & 123.6 & 0.1 &  & 123 & 123 & \textbf{123} & 22.4  \\
    20 & 15 & 200 & 168 & 30.4 &  & 168 & 170 & 169 & 15.5 &  & 164 & 165 & 164.5 & 0.3 &  & 164 & 164 & \textbf{164} & 30.2  \\
    20 & 15 & 250 & 214 & 30.5 &  & 212 & 214 & 213.2 & 21.7 &  & 206 & 207 & 206.6 & 0.4 &  & 206 & 206 & \textbf{206} & 25.6  \\
    20 & 15 & 300 & 266 & 30.6 &  & 253 & 257 & 254.3 & 29.9 &  & 246 & 247 & 246.3 & 0.5 &  & 246 & 246 & \textbf{246} & 38.6  \\
    20 & 15 & 350 & 308 & 30.7 &  & 294 & 297 & 296 & 38.5 &  & 289 & 290 & 289.2 & 1.1 &  & 288 & 288 & \textbf{288} & 34.2  \\
    20 & 15 & 400 & 340 & 60.8 &  & 336 & 337 & 336.3 & 48.3 &  & 328 & 328 & 328 & 1.2 &  & 327 & 328 & \textbf{327.2} & 80.8  \\
    20 & 15 & 800 & 711 & 61.7 &  & 665 & 669 & 667.5 & 170.3 &  & 655 & 656 & 655.2 & 5.7 &  & 655 & 655 & \textbf{655} & 108.1  \\
    20 & 15 & 1000 & 883 & 122.1 &  & 829 & 833 & 830.7 & 254.6 &  & 821 & 821 & 821 & 8.6 &  & 820 & 820 & \textbf{820} & 118.1  \\
    20 & 15 & 1500 & 1335 & 123.9 &  & 1246 & 1251 & 1247.9 & 545.5 &  & 1228 & 1229 & \textbf{1228.2} & 22.3 &  & 1229 & 1229 & 1229 & 140.1  \\
    20 & 20 & 50 & 45 & 304 &  & 46 & 46 & 46 & 1.1 &  & 43 & 44 & \textbf{43.8} & 0.01 &  & 44 & 44 & 44 & 22.3  \\
    20 & 20 & 100 & 90 & 30.3 &  & 88 & 90 & 89.1 & 6.2 &  & 86 & 86 & \textbf{86} & 0.06 &  & 87 & 87 & 87 & 21  \\
    20 & 20 & 150 & 134 & 30.5 &  & 132 & 136 & 134.2 & 3.3 &  & 127 & 128 & 127.1 & 0.1 &  & 127 & 127 & \textbf{127} & 17.1  \\
    20 & 20 & 200 & 181 & 30.5 &  & 174 & 177 & 174.7 & 16.9 &  & 169 & 170 & 169.5 & 0.2 &  & 168 & 168 & \textbf{168} & 21.2  \\
    20 & 20 & 250 & 227 & 31.4 &  & 217 & 219 & 217.8 & 23.5 &  & 211 & 212 & 211.4 & 0.6 &  & 211 & 211 & \textbf{211} & 22.4  \\
    20 & 20 & 300 & 271 & 30.8 &  & 257 & 259 & 257.8 & 32.2 &  & 253 & 254 & 253.1 & 0.6 &  & 253 & 253 & \textbf{253} & 27.6  \\
    20 & 20 & 350 & 303 & 31 &  & 302 & 304 & 302.7 & 41.2 &  & 295 & 296 & 295.1 & 1.1 &  & 295 & 295 & \textbf{295} & 30.2  \\
    20 & 20 & 400 & 392 & 61.2 &  & 344 & 347 & 345.4 & 51.4 &  & 337 & 338 & 337.2 & 1.3 &  & 337 & 337 & \textbf{337} & 54.7  \\
    20 & 20 & 800 & 739 & 62.2 &  & 687 & 689 & 688.2 & 183.2 &  & 674 & 675 & 674.5 & 6.1 &  & 673 & 673 & \textbf{673} & 58.8  \\
    20 & 20 & 1000 & 942 & 125.4 &  & 850 & 856 & 852.1 & 270.2 &  & 839 & 840 & 839.2 & 10.1 &  & 839 & 839 & \textbf{839} & 105  \\
    20 & 20 & 1500 & 1429 & 124.3 &  & 1271 & 1274 & 1272.4 & 569.5 &  & 1257 & 1259 & \textbf{1258} & 24.6 &  & 1258 & 1259 & 1258.2 & 114.2  \\
    20 & 25 & 50 & 46 & 30.3 &  & 46 & 47 & 46.3 & 1.3 &  & 45 & 45 & \textbf{45} & 0.02 &  & 45 & 45 & \textbf{45} & 31.9  \\
    20 & 25 & 100 & 91 & 30.5 &  & 89 & 91 & 90.3 & 7 &  & 87 & 87 & 87 & 0.07 &  & 86 & 86 & \textbf{86} & 15  \\
    20 & 25 & 150 & 137 & 30.5 &  & 136 & 138 & 137.5 & 12.4 &  & 130 & 131 & 130.2 & 0.1 &  & 130 & 130 & \textbf{130} & 21.1  \\
    20 & 25 & 200 & 180 & 30.8 &  & 176 & 182 & 179.9 & 18.1 &  & 172 & 173 & 172.4 & 0.3 &  & 172 & 173 & \textbf{172.2} & 24.2  \\
    20 & 25 & 250 & 234 & 30.8 &  & 221 & 227 & 223.2 & 25.2 &  & 215 & 216 & 215.1 & 0.5 &  & 215 & 215 & \textbf{215} & 25.3  \\
    20 & 25 & 300 & 274 & 31 &  & 265 & 266 & 265.7 & 34.2 &  & 258 & 259 & 258.5 & 0.7 &  & 258 & 258 & \textbf{258} & 37.8  \\
    20 & 25 & 350 & 335 & 31.1 &  & 309 & 310 & 309.4 & 43.9 &  & 300 & 301 & 300.6 & 1.1 &  & 300 & 301 & \textbf{300.2} & 29.2  \\
    20 & 25 & 400 & 373 & 61.4 &  & 350 & 352 & 351.1 & 54.3 &  & 343 & 344 & 343.7 & 1.4 &  & 343 & 343 & \textbf{343} & 59.1  \\
    20 & 25 & 800 & 801 & 65.2 &  & 669 & 703 & 701.1 & 191.2 &  & 684 & 685 & \textbf{684.6} & 6.7 &  & 684 & 685 & \textbf{684.5} & 66  \\
    20 & 25 & 1000 & 1001 & 126.3 &  & 863 & 864 & 863.8 & 284 &  & 853 & 853 & \textbf{853} & 11.1 &  & 853 & 853 & \textbf{853} & 112.6  \\
    20 & 25 & 1500 & 1425 & 125.2 &  & 1294 & 1299 & 1296.9 & 589.9 &  & 1280 & 1281 & \textbf{1280.3} & 25.8 &  & 1280 & 1281 & 1280.3 & 142.6  \\       
20 & 40 & 50 & 48 & 30.6 &  & 48 & 49 & 48.7 & 2.1 &  & 46 & 46 & \textbf{46} & 0.02 &  & 46 & 46 & \textbf{46} & 18.3  \\
    20 & 40 & 100 & 92 & 30.5 &  & 92 & 93 & 92.7 & 9.1 &  & 90 & 90 & \textbf{90} & 0.09 &  & 90 & 90 & \textbf{90} & 24.9  \\
    20 & 40 & 150 & 140 & 30.8 &  & 137 & 140 & 138.8 & 6.9 &  & 133 & 134 & \textbf{133.8} & 0.2 &  & 135 & 135 & 135 & 23.5 \\
    20 & 40 & 200 & 180 & 31 &  & 181 & 183 & 182.4 & 22.5 &  & 178 & 178 & \textbf{178} & 0.4 &  & 178 & 178 & \textbf{178} & 32.9  \\
    20 & 40 & 250 & 233 & 31.3 &  & 225 & 226 & 225.1 & 30.6 &  & 221 & 222 & \textbf{221.7} & 0.5 &  & 222 & 222 & 222 & 25.3  \\
    20 & 40 & 300 & 276 & 31.6 &  & 272 & 273 & 272.5 & 40.8 &  & 265 & 267 & \textbf{266.1} & 1.1 &  & 266 & 267 & 266.8 & 37.7  \\
    20 & 40 & 350 & 321 & 31.8 &  & 320 & 327 & 321.4 & 51.9 &  & 309 & 310 & \textbf{309.3} & 1.5 &  & 310 & 310 & 310 & 34.8  \\
    20 & 40 & 400 & 390 & 62.2 &  & 361 & 363 & 361.6 & 63.6 &  & 352 & 353 & \textbf{352.3} & 2 &  & 353 & 353 & 353 & 56.9  \\
    20 & 40 & 800 & 801 & 64.5 &  & 711 & 714 & 712 & 219.7 &  & 703 & 704 & 703.4 & 8.2 &  & 702 & 702 & \textbf{702} & 70.9  \\
    20 & 40 & 1000 & 1001 & 137.4 &  & 895 & 899 & 896.6 & 315.5 &  & 877 & 879 & \textbf{878.4} & 13.4 &  & 879 & 880 & 879.2 & 114.7  \\
    20 & 40 & 1500 & 1501 & 128.3 &  & 1330 & 1333 & 1331.4 & 625.8 &  & 1316 & 1317 & 1316.9 & 30.5 &  & 1316 & 1317 & \textbf{1316.4} & 129.9  \\
    20 & 60 & 50 & 47 & 30.8 &  & 48 & 48 & 48 & 3.4 &  & 46 & 47 & 46.4 & 0.03 &  & 46 & 46 & \textbf{46} & 19.4  \\
    20 & 60 & 100 & 94 & 30.8 &  & 97 & 97 & 97 & 6 &  & 92 & 92 & \textbf{92} & 0.2 &  & 92 & 92 & \textbf{92} & 18.7  \\
    20 & 60 & 150 & 139 & 31.2 &  & 141 & 142 & 141.9 & 10.9 &  & 137 & 138 & 137.1 & 0.2 &  & 137 & 137 & \textbf{137} & 31.2  \\
    20 & 60 & 200 & 185 & 31.7 &  & 186 & 187 & 186.2 & 28.2 &  & 181 & 182 & \textbf{181.7} & 0.4 &  & 182 & 182 & 182 & 32.2  \\
    20 & 60 & 250 & 239 & 32.1 &  & 230 & 234 & 232.3 & 37.5 &  & 226 & 227 & \textbf{226.2} & 0.9 &  & 227 & 227 & 227 & 33.6  \\
    20 & 60 & 300 & 301 & 32.5 &  & 274 & 277 & 275.2 & 49.5 &  & 270 & 270 & \textbf{270} & 1.2 &  & 271 & 271 & 271 & 31.2  \\
    20 & 60 & 350 & 321 & 32.9 &  & 319 & 320 & 319.9 & 61.8 &  & 314 & 315 & \textbf{314.4} & 1.6 &  & 315 & 315 & 315 & 37.3  \\
    20 & 60 & 400 & 381 & 63.4 &  & 365 & 368 & 366.5 & 75.8 &  & 359 & 359 & \textbf{359} & 2.3 &  & 361 & 361 & 361 & 86.7  \\
    20 & 60 & 800 & 801 & 70.2 &  & 724 & 727 & 725.3 & 252.1 &  & 715 & 717 & \textbf{715.5} & 9.9 &  & 716 & 717 & 716.1 & 68.1  \\
    20 & 60 & 1000 & 1001 & 128.4 &  & 910 & 914 & 911.2 & 355.9 &  & 896 & 897 & 896.2 & 16.4 &  & 896 & 896 & \textbf{896} & 121.8  \\
    20 & 60 & 1500 & 1501 & 132.9 &  & 1356 & 1359 & 1357.5 & 700.1 &  & 1340 & 1341 & 1340.6 & 38.2 &  & 1340 & 1341 & \textbf{1340.2} & 131.6  \\
    20 & 80 & 50 & 48 & 30.6 &  & 49 & 49 & 49 & 4 &  & 47 & 48 & 47.1 & 0.07 &  & 47 & 47 & \textbf{47} & 22.5  \\
    20 & 80 & 100 & 95 & 31.2 &  & 97 & 97 & 97 & 8 &  & 93 & 93 & \textbf{93} & 0.1 &  & 93 & 93 & \textbf{93} & 28.8  \\
    20 & 80 & 150 & 140 & 31.9 &  & 141 & 142 & 141.9 & 24.1 &  & 138 & 139 & 138.1 & 0.3 &  & 138 & 138 & \textbf{138} & 29.4  \\
    20 & 80 & 200 & 190 & 32.3 &  & 187 & 188 & 187.1 & 34 &  & 183 & 184 & 183.5 & 0.6 &  & 183 & 183 & \textbf{183} & 34.2  \\
    20 & 80 & 250 & 234 & 32.5 &  & 233 & 239 & 236 & 20.8 &  & 229 & 230 & 229.1 & 0.9 &  & 229 & 229 & \textbf{229} & 39.1  \\
    20 & 80 & 300 & 283 & 35.5 &  & 280 & 283 & 281.9 & 58.2 &  & 274 & 275 & 274.3 & 1.5 &  & 274 & 275 & \textbf{274.1} & 38.3  \\
    20 & 80 & 350 & 330 & 37.5 &  & 327 & 331 & 328.6 & 72.4 &  & 318 & 320 & \textbf{319} & 2.1 &  & 319 & 319 & \textbf{319} & 39.9  \\
    20 & 80 & 400 & 373 & 68.1 &  & 369 & 373 & 371.1 & 87.8 &  & 364 & 364 & \textbf{364} & 2.8 &  & 364 & 365 & 364.2 & 66.9  \\
    20 & 80 & 800 & 801 & 69.4 &  & 739 & 742 & 740.7 & 288 &  & 724 & 726 & \textbf{725.2} & 12.3 &  & 726 & 727 & 726.4 & 83.5  \\
    20 & 80 & 1000 & 1001 & 131.4 &  & 921 & 924 & 922.5 & 399.9 &  & 904 & 905 & \textbf{904.5} & 19.6 &  & 908 & 908 & 908 & 142.1  \\
    20 & 80 & 1500 & 1501 & 137.3 &  & 1367 & 1370 & 1368.3 & 772.1 &  & 1354 & 1355 & \textbf{1354.3} & 44.3 &  & 1354 & 1355 & 1354.4 & 144.7  \\
    20 & 100 & 50 & 48 & 30.5 &  & 49 & 49 & 49 & 5.1 &  & 48 & 48 & \textbf{48} & 0.09 &  & 48 & 48 & \textbf{48} & 20.6  \\
    20 & 100 & 100 & 97 & 32.7 &  & 96 & 97 & 96.9 & 9.9 &  & 93 & 94 & \textbf{93.8} & 0.1 &  & 93 & 94 & \textbf{93.8} & 32.1  \\
    20 & 100 & 150 & 144 & 31.8 &  & 141 & 142 & 141.2 & 15.7 &  & 139 & 139 & 139 & 0.3 &  & 139 & 139 & \textbf{139} & 27.9  \\
    20 & 100 & 200 & 190 & 32.1 &  & 188 & 191 & 190.4 & 39.5 &  & 185 & 185 & 185 & 0.8 &  & 185 & 185 & \textbf{185} & 37.1  \\
    20 & 100 & 250 & 236 & 36.5 &  & 234 & 236 & 235.1 & 25.3 &  & 230 & 231 & 230.5 & 1.1 &  & 230 & 230 & \textbf{230} & 37.2  \\
    20 & 100 & 300 & 283 & 35.1 &  & 280 & 282 & 280.9 & 67.1 &  & 276 & 276 & \textbf{276} & 1.6 &  & 276 & 276 & \textbf{276} & 39.9  \\
    20 & 100 & 350 & 333 & 39.5 &  & 328 & 333 & 330 & 82.9 &  & 322 & 322 & \textbf{322} & 2.4 &  & 322 & 322 & \textbf{322} & 40.6  \\
    20 & 100 & 400 & 375 & 70.9 &  & 372 & 373 & 372.2 & 100 &  & 367 & 367 & \textbf{367} & 3.2 &  & 367 & 367 & \textbf{367} & 72.6  \\
    20 & 100 & 800 & 801 & 71.5 &  & 743 & 745 & 744 & 321.3 &  & 730 & 731 & \textbf{730.2} & 14.6 &  & 730 & 731 & \textbf{730.2} & 76.8  \\
    20 & 100 & 1000 & 1001 & 146.8 &  & 923 & 925 & 924 & 442.9 &  & 911 & 912 & \textbf{911.8} & 23.1 &  & 913 & 913 & 913 & 138.3  \\
    20 & 100 & 1500 & 1501 & 143.7 &  & 1381 & 1384 & 1382.1 & 826.1 &  & 1364 & 1365 & \textbf{1364.2} & 52.2 &  & 1366 & 1367 & 1366.6 & 160.5  \\
\end{longtable}}



\end{document}